\documentclass[10pt,journal,compsoc]{IEEEtran}
\newif\ifpeerreview

\peerreviewfalse

\usepackage[nocompress]{cite}
\usepackage{url}
\usepackage{amsmath,amssymb,graphicx}

\usepackage{lipsum} 

\usepackage[switch]{lineno}

\usepackage{tabularx}
\usepackage{booktabs}
\usepackage{multirow}
\usepackage{bbm}
\usepackage{amsmath}

\newcommand{\myparagraph}[1]{\vspace{0.1em}\noindent\textbf{#1}}

\newcommand{\paperID}{17}

\title{Convolutional Neural Opacity Radiance Fields}

\author{Haimin~Luo, Anpei~Chen, Qixuan~Zhang, Bai~Pang, Minye~Wu, Lan~Xu, 
        and~Jingyi~Yu,~\IEEEmembership{Fellow,~IEEE}
\IEEEcompsocitemizethanks{
\IEEEcompsocthanksitem H. Luo, Q. Zhang, B. Pang, L. Xu and J. Yu are with the School of Information Science and Technology, ShanghaiTech University, Shanghai 201210, China. E-mail:\{luohm, zhangqx1, pangbai, xulan1, yujingyi\}@shanghaitech.edu.cn.

\IEEEcompsocthanksitem A. Chen and M. Wu are with the School of Information Science and Technology,
ShanghaiTech University, Shanghai 201210, China, and the Shanghai
Institute of Microsystem and Information Technology, Chinese Academy of
Sciences, Shanghai 200031, China, and also with the University of Chinese
Academy of Sciences, Beijing 100049, China.\\ E-mail:\{chenap, wumy\}@shanghaitech.edu.cn.

\IEEEcompsocthanksitem J. Yu is with the Shanghai Engineering Research Center of Intelligent Vision and Imaging, School of Information Science and Technology, ShanghaiTech University, Shanghai 201210, China. E-mail: yujingyi@shanghaitech.edu.cn.
}
}

\begin{document}

\IEEEtitleabstractindextext{%
\begin{abstract}
Photo-realistic modeling and rendering of fuzzy objects with complex opacity are critical for numerous immersive VR/AR applications, but it suffers from strong view-dependent brightness, color. In this paper, we propose a novel scheme to generate opacity radiance fields with a convolutional neural renderer for fuzzy objects, which is the first to combine both explicit opacity supervision and convolutional mechanism into the neural radiance field framework so as to enable high-quality appearance and global consistent alpha mattes generation in arbitrary novel views. More specifically, we propose an efficient sampling strategy along with both the camera rays and image plane, which enables efficient radiance field sampling and learning in a patch-wise manner, as well as a novel volumetric feature integration scheme that generates per-patch hybrid feature embeddings to reconstruct the view-consistent fine-detailed appearance and opacity output. We further adopt a patch-wise adversarial training scheme to preserve both high-frequency appearance and opacity details in a self-supervised framework. We also introduce an effective multi-view image capture system to capture high-quality color and alpha maps for challenging fuzzy objects. Extensive experiments on existing and our new challenging fuzzy object dataset demonstrate that our method achieves photo-realistic, globally consistent, and fined detailed appearance and opacity free-viewpoint rendering for various fuzzy objects.
\end{abstract}

\begin{IEEEkeywords} 
Computational Photography, Neural Rendering, Opacity Modelling, View Synthesis.
\end{IEEEkeywords}
}

\maketitle
\thispagestyle{empty}


\IEEEraisesectionheading{
  \section{Introduction}\label{sec:introduction}
}


\begin{figure*}[!t]
\centering
    \includegraphics[width=1.0\linewidth]{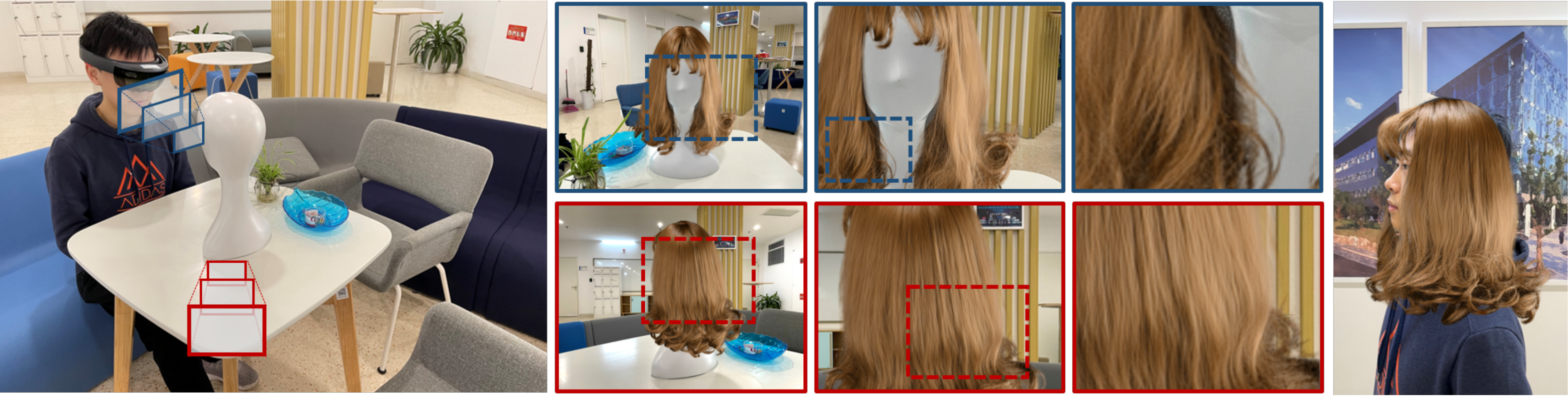}
\caption{VR$/$AR experience with photo-realistic opacity radiance field rendering (first column), our method is able to reconstruct fuzzy appearance at different scales (2 - 5 columns).}
\label{fig:teaser}
\end{figure*}

\IEEEPARstart{T}{he} past ten years have witnessed a rapid development of 3D reconstruction technologies for complex scenes with the popularity of commercial passive and active image sensors, which enables numerous immersive experience and virtual and augmented reality (VR and AR) applications and
has recently attracted substantive attention.
However, the photo-realistic modeling of fuzzy objects with complex opacity such as hair, fur and feathers remains unsolved, which suffers from strong view-dependent brightness, color changes, leading to difficulties in both geometry and appearance reconstruction.


For high-quality fuzzy object modeling, early solutions~\cite{paris2008hair, luo2013structure, DynHair_2014, Strand_CVPR2019} require costly capture devices and systems, coded lighting, or even manually effort to achieve high-fidelity hair strand geometry reconstruction, which is difficult to be deployed for daily usage.
%
To avoid the heavy reliance on precise geometry modeling, researchers adopt image-based rendering (IBR)~\cite{Buehler2001Unstructured,matusik2002image, LFView_2016} to reconstruct the appearance of furry objects by interpolating new views from the captured ones.
Specifically, to handle opacity objects, the traditional approach~\cite{matusik2002image} utilizes multi-view images and alpha mattes to compute the angular opacity maps in novel views, which suffers from severe ghosting effect caused by insufficient view samples and inaccurate geometry proxies.
Moreover, obtaining accurate geometry of furry objects is intractable since fur and hair contain tens of thousands of thin fibers and their mutual occlusions are the fundamental causes of translucency.

Only recently, the neural rendering techniques~\cite{sitzmann2019srns,lombardi2019neural,thies2019deferred,wang2020neural,mildenhall2020nerf} bring huge potential for photo-realistic novel view synthesis from only images input, with various data representations such as point-clouds~\cite{aliev2019neural,wang2020neural}, voxels~\cite{lombardi2019neural,sitzmann2019deepvoxels}, meshes~\cite{NeuralMesh_CVPR2018,thies2019deferred} or implicit representation~\cite{park2019deepsdf,mescheder2019occupancy,sitzmann2019srns,mildenhall2020nerf}.
However, the literature on fuzzy object neural rendering remains sparse.
The recent approach~\cite{wang2020neural} utilizes a neural point renderer to generate texture maps and alpha mattes in novel views explicitly. 
However, extracting features from a coarse point cloud leads to insufficient sampling and severe artifacts when zooming in and out.
Besides, researchers ~\cite{mildenhall2020nerf,liu2020neural,martin2020nerf,park2020deformable} combine implicit representation with volume rendering to inherently model the density of the continuous 3D space, achieving state-of-the-art appearance rendering results, even for opacity objects.
%
Though these methods achieve view-consistent fuzzy object modeling, their method still lacks fine details in both texture and alpha.
Specifically, only using a Multi-Layer Perceptron (MLP) network is adopted to fit the continuous 3D space, leading to uncanny high-frequency appearance and opacity details. Most recently, Positional encoding~\cite{mildenhall2020nerf} and Fourier feature mapping~\cite{tancik2020fourier} schemes enable the MLP to handle richer texture details, but it still fails to recover fuzzy surface which contains extremely high-frequency variation.


In this paper, we attack the above challenges and propose a novel scheme to generate convolutional neural opacity radiance fields for fuzzy objects, which is the first to combine explicit opacity supervision with neural radiance field technique (See Fig.~\ref{fig:teaser} for an overview). 
Our novel pipeline enables high-quality appearance and global consistent alpha mattes generation in arbitrary novel views for more immersive VR/AR applications.

More specifically, to provide explicit opacity supervision, from the system side we introduce an effective multi-view capture system equipped with a step turntable and specific transparent lighting design as well as a corresponding multi-color keying algorithm.
Our novel system automatically captures both high-quality RGB images and corresponding opacity maps for challenging fuzzy objects in the input capture views. 
Based on such hybrid input, from the algorithm side, we introduce the convolutional mechanism in the image plane into the neural radiance field framework~\cite{martin2020nerf} so that enable photo-realistic appearance and global consistent opacity generation in arbitrary novel views. 
To this end, we first propose an efficient sampling strategy that utilizes the inherent silhouette prior along the rays and encodes the spatial information across the image plane, which enables efficient radiance field sampling and learning in a patch-wise manner.
Then, we perform a novel volumetric integration scheme to generate a per-patch hybrid appearance and opacity feature maps, followed by a light-weight convolutional U-Net to reconstruct the view-consistent fine-detailed appearance and opacity output.
Moreover, a patch-wise adversarial training scheme is proposed to preserve both high-frequency appearance and opacity details for photo-realistic rendering in a self-supervised framework.
%
To summarize, our main contributions include:
\begin{itemize}
    \item We present a novel convolutional neural radiance field generation scheme to reconstruct high frequency and global consistent appearance and opacity of fuzzy objects in novel views, achieving significant superiority to the existing state of the art.
    
    \item To enable convolutional mechanism, we propose an efficient sampling strategy, a hybrid feature integration as well as a self-supervised adversarial training scheme for patch-wise radiance field learning. 
    
    \item We introduce an effective multi-view system to capture the color and alpha maps for challenging fuzzy objects, and our capture dataset will be made available to stimulate further research.
\end{itemize}

\begin{figure*}[!ht]
	\centering
	\includegraphics[width=1.0\linewidth]{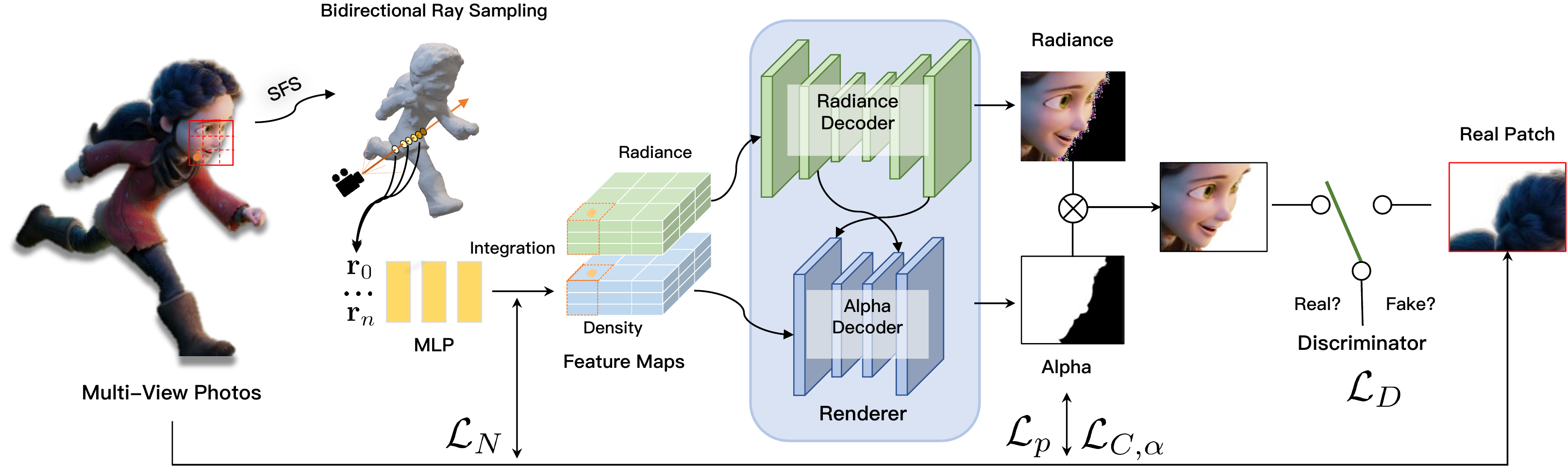}
	\caption{Overview of our end-to-end ConvNeRF pipeline. Given multi-view RGBA images, we use an SFS to infer proxy geometric for Efficient Ray Sampling. For each sample point in the volume space, the position and direction are feeding to an MLP based feature prediction network to represent the object at a global level. We next concatenate nearby rays into local feature patches and decoded them into RGB and matte with the convolutional volume renderer. An adversarial training strategy is used on the final output to encourage fine surface details. In the reference period, we render the entire image at once rather than per patch rendering. }
	\label{fig:pipeline}
\end{figure*}

\section{Related Work}

\subsection{Neural 3D Shape Modeling}
Recent work has made a significant process on 3D object modeling and realism free-viewpoint rendering with level sets of deep networks that implicitly map spacial locations $xyz$ to a geometric representation (i.g., distance field ~\cite{park2019deepsdf}, occupancy Field ~\cite{mescheder2019occupancy,chen2019learning} etc.).  The aforementioned explicit representations require discretization (e.g., in terms of the number of voxels, points or vertices), implicitly models shapes with a continuous function and naturally is able to handle complicated shape topologies. The implicit geometric modeling can be quickly learned from 3D point samples \cite{peng2020convolutional,saito2019pifu}, and the trained models can be used to reconstruct shapes from a single image or 3D part. However, these models are limited by their requirement of access to ground truth 3D geometry, typically obtained from synthetic 3D shape datasets such as ShapeNet~\cite{chang2015shapenet}. Subsequent works relax this requirement by formulating differentiable rendering functions that allow neural implicit shape representations to be optimized using only 2D images ~\cite{sitzmann2019srns,niemeyer2020differentiable}. 


\subsection{Free-Viewpoint Rendering}
Free-viewpoint synthesis methods are generally model input/target images as a collection of rays and essentially aims to recover the plenoptic function ~\cite{Debevec1998Image} from the dense samples. Earlier Image-Based Rendering (IBR) work~\cite{Levoy1996Light} used two planes ($uvst$) parametrization or 2PP to represent rays and render new rays via a weighted blending of the ray samples, i.e., fusing nearby rays by considering view angle and camera spacial distance. They are able to achieve real-time interpolation but require much memory as they need to cache all rays. Following work ~\cite{Buehler2001Unstructured} bring in proxy geometric to select suitable views and filter occluded rays by cross-projection to the image plane when ray fusion. 
However, those methods are still limited by the linear blending function, leading to severe ghosting and blurring artifact. 

Most recently, seminal researches seek to implicitly represent the radiance field and render novel views with a neural network. Deep Surface Light Fields~\cite{chen2018deep} use an MLP network to fix per-vertex radiance and learns to fill up the missing data across angles  and vertices. Deferred neural rendering~\cite{thies2019deferred} presents a novel learnable neural texture to model rendering as image translation, which uses a coarse geometry for texture projection and offers flexible content editing. Recent works~\cite{sitzmann2019deepvoxels, martin2020nerf, zhang2020nerf++} present a learned representation that encodes the view-dependent appearance of a 3D scene without modeling its geometry explicitly. Another line of research extends the free-viewpoint to animation squeezing or scene relighting by modeling and rendering dynamic scenes through embedding spacial feature with sparse dynamic point cloud~\cite{Wu_CVPR2020}, using volumetric representation to reconstruct dynamic geometry and appearance variations jointly with only image-level supervision~\cite{lombardi2019neural}, or modeling image formation in terms of environment lighting, object intrinsic attributes and the light transport function~\cite{chen2020neural}. A Notable exception is NeRF~\cite{mildenhall2020nerf}, which implicitly models the radiance field and the density of a volume with a neural network, then uses a direct volume rendering function to synthesize novel views. They also demonstrate a heretofore unprecedented level of fidelity on a range of challenging scenes. The following work NeRF-W~\cite{martin2020nerf} relaxes the NeRF's strict consistency assumptions through modeling per-image appearance variations such as exposure, lighting, weather, and post-processing with a learned low-dimensional latent space. However, such pure ray-based rendering schemes are failing to recover high-frequency surface, such as fur.

\subsection{Image Matting}
Traditional natural image matting algorithms usually rely on user-defined trimap~\cite{chuang2001bayesian} or scribble~\cite{wang2005iterative} as additional input and can generally be divided into sampling-based methods and propagation-based methods. 
Sampling based methods~\cite{gastal2010shared,shahrian2013improving,karacan2015image,elhamifar2015dissimilarity}, the known foreground and background regions are sampled as candidates to find the best pixel pair by a carefully designed metric for estimating the alpha value of a query pixel in the unknown area. 
%
%
In propagation-based methods, the alpha values in unknown regions are propagated from the known foreground and background regions using a reformulated image matting model according to different affinities between pixels, using various propagation schemes~\cite{sun2004poisson,grady2005random,levin2007closed,chen2013knn,chen2013image}.


Recently, learning-based methods have made an impressive process. Deep Image Matting~\cite{xu2017deep} builds a large matting dataset by alpha composition and proposes an end-to-end deep learning framework to automatically compute alpha mattes using RGB image and trimap as input. Following this work, different methods have been proposed. 
Disentangled Image Matting~\cite{cai2019disentangled} adapts the input trimap as well as estimates alpha with a novel multi-task loss. HDMatting~\cite{yu2020high} estimate the alpha matte of high resolution images patch by patch through a Cross-Patch Contextual module guided by the given trimap. Recent methods not only estimate the alpha matte but the foreground image~\cite{hou2019context} and the background image~\cite{forte2020f} such that the popular perceptual loss~\cite{johnson2016perceptual, hou2019context} or a fusion mechanism~\cite{forte2020f} can be adopted. To achieve trimap-free matting, recent works implicitly generate trimap using a late fusion model with a soft segmentation network~\cite{zhang2019late} or require additional background photos as input~\cite{sengupta2020background} for self-supervised training.

 
Rather than recover alpha from a 2D image, we set out to generate free-viewpoint alpha matte directly following 3D constraints.

\section{Algorithm Details}
\label{sec: algorithm}

In this section, we introduce the design of our convolutional neural opacity radiance fields (ConvNeRF in short), which enables photo-realistic global-consistent appearance and opacity rendering in novel views based on the RGBA input from our capture system, as illustrated in Fig.~\ref{fig:pipeline}.

Our key insight is to encode opacity information explicitly with a spatial convolutional mechanism to enhance the neural radiance field approach NeRF~\cite{mildenhall2020nerf} for high-frequency detail modeling.
Inspired by NeRF, we adopt the similar implicit neural radiance field to represent a scene using a Multi-Layer Perceptron (MLP), as well as the volumetric integration of predicted density and color values along the casting rays.
Please refer to NeRF~\cite{mildenhall2020nerf} for more details. 

Differently, our ConvNeRF further encodes opacity explicitly with spatial convolutional design to significantly improve the neural radiance field reconstruction.
To this end, we first propose an efficient sampling strategy to not only utilize the inherent silhouette prior along the camera rays but also encode the spatial information across the image plane (Sec.~\ref{sec:BSampling}). 
%
Then, a global geometric representor is adopted to map a 3D location to a high-level radiance feature and then a novel volumetric integration scheme is adopted to generate per-patch hybrid feature embeddings, which models the characteristics of appearance and opacity separately for more effective radiance field learning in a patch-wise manner (Sec.~\ref{sec:Feature_patch}).
%
Next, we choose a light-weight U-Net to decode the feature patches into view-consistent appearance and opacity output (Sec.~\ref{sec:Convolutional Volume Renderer}).
%
We further adopt a patch-wise adversarial training scheme to preserve both high-frequency appearance and opacity details in a self-supervised framework (Sec.\ref{sec:Training}).



\begin{figure}[!t]
	\centering
	\includegraphics[width=1.0\linewidth]{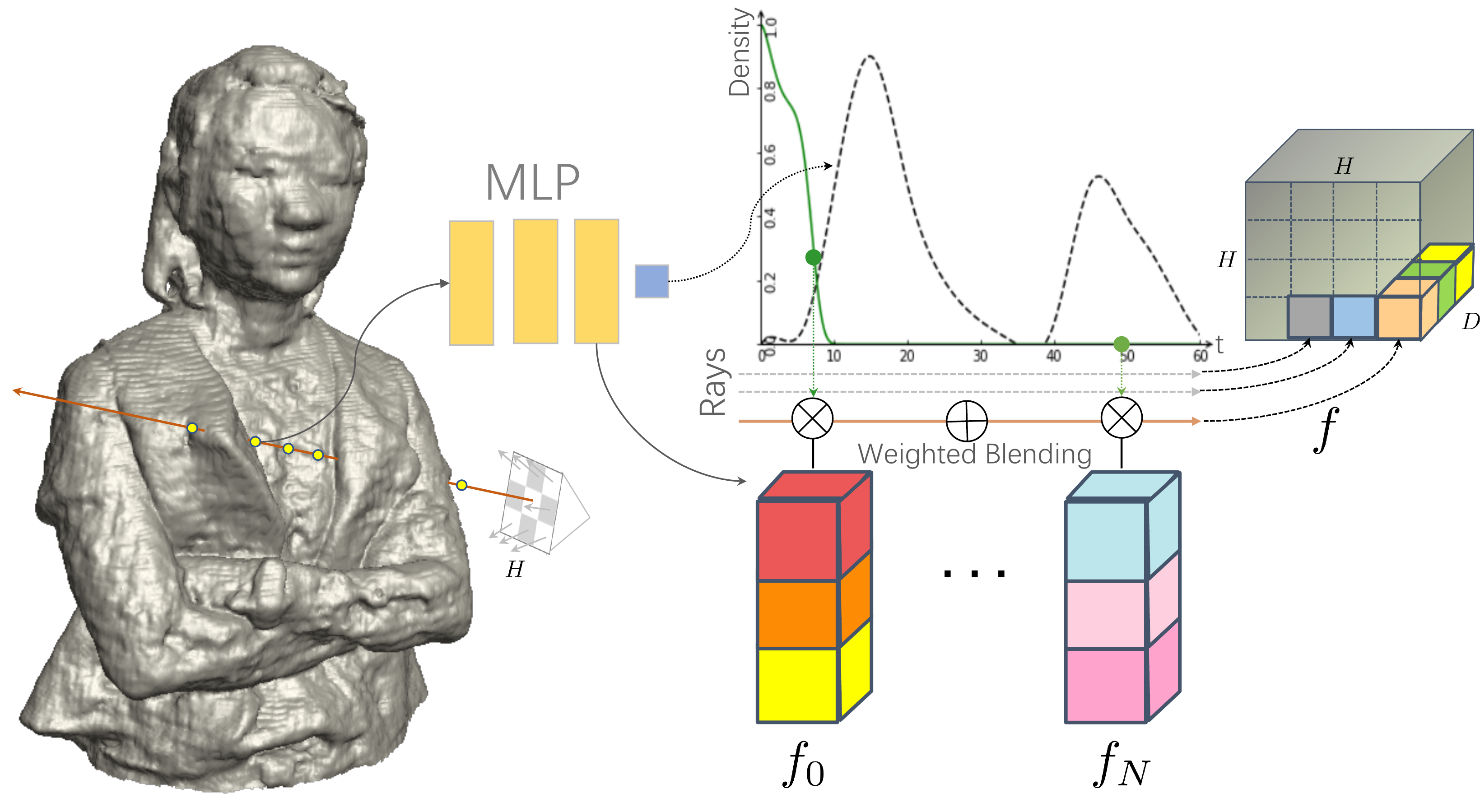}
	\caption{Feature patches construction. To enable 2D convolution, we weighted blend the volume features of each ray as the ray feature with the "cumproded"  density of the sample points and concatenate nearby ray features into local patches for the following convolutional rendering.}
	\label{fig:feature}
\end{figure}

\subsection{Efficient Ray Sampling} \label{sec:BSampling}

To speed up the training procedure, we introduce a patch-wise sampling scheme, which uses a coarse proxy to filter out redundant samples for more efficient radiance field sampling and rendering.


%

\myparagraph{Coarse Proxy Generation.} 
Note that the multi-view alpha mattes encodes the silhouette prior of the captured fuzzy object inherently, which provides a reliable proxy to guide the sampling process in the continuous 3D space. 
To this end, the input alpha mattes are binarized and dilated, and the Shape-from-Silhouette (SfS)~\cite{SFS} algorithm is applied to obtain a coarse 3D proxy of the fuzzy object. 

\myparagraph{Spatial Sampling.}
To further enable patch-wise spatial sampling across the image plane, we first render two depth maps: the near-depth ( the first hit point) and far-depth (the last hit point)  for all training views by projecting the proxy mesh to the image plane. Then, we uniformly divide the input images and depth maps into  $K \times K$ small patches without spacing and filter out the patches outside the valid near/far-depth (i.e., pure background patches). 
%

\myparagraph{Ray-wise Sampling.}
Given the above sampled ray patches, we perform a similar two-stage sampling scheme of the original NeRF~\cite{mildenhall2020nerf} in a coarse-to-fine manner for ray-wise radiance fields sampling to sample $N$ sampling points for each ray of the ray patches. Differently, note that we have obtained near-far depth priors for sampled rays, we perform the ray-wise sampling on the near-far region only in both coarse and fine stage. Thus it requires only $1/8$ sample points of NeRF at least and achieves much more efficient radiance field training and rendering.

The sampled patches with tightly sampling points in valid regions are then utilized for the following training process of the neural radiance field in a patch-wise manner. In practice, to balance the memory requirement and sampling effectiveness, during training $K$ is set to be 32 and we use 12 patches as a batch with 64 samples per ray. It achieves about 4 times faster training and about 10 times faster inference compared to the original NeRF~\cite{mildenhall2020nerf} in general.

\subsection{Patch-wise integration}\label{sec:Feature_patch}

%
Unlike NeRF predicting pixel color by integrating in a low-dimensional color space, we instead perform a novel volumetric integration scheme to generate a per-patch hybrid appearance and opacity feature maps to enable more effective radiance field learning in a patch-wise manner.   

In our ConvNeRF, for a given 3D location $\textbf{x}$ and viewing direction $\bf d$, we first extract its 3D features $(\textbf{f}, \sigma)$ with a global geometric representor  $\bf E_\Theta$:
\begin{equation}
    (\textbf{f}, \sigma) = \bf E_{\Theta}(\bf x, \bf d).
\end{equation}
Same as the original NeRF network architecture, $\bf E_\Theta$ is achieved with an MLP block with optimizable weights $\bf \Theta$, we remove the last layer of RGB color branch so that $\bf E_\Theta$ plays as a global feature extractor and generates both radiance term $\bf f$ and the geometric term $\sigma$ output. 


Then for each ray of the patches, we sample $N$ points and extract their 3D features to predict the foreground and background probability (i.e., the alpha):

\begin{equation}
    \alpha_i = T_i(1 - exp(-\sigma_i\delta_i)), T_i=exp(-\sum_{j=1}^{i-1}\sigma_j\delta_j),
\label{eq:alpha}
\end{equation}
where $\delta$ is the distance between adjacent sampling points.
The projected feature for each patch pixel is obtained by an integration scheme alone the ray:
\begin{equation}
    F_c = \sum_{i=1}^{N}\alpha_i \bf f_i,
\label{eq:integration}
\end{equation}

As mentioned above, we decompose the free-viewpoint object rendering into radiance and alpha components; thus, we expect to decouple those two components when rendering. 
To this end, different from radiance feature extraction which fuses the radiance term of each sampling point along a ray by integration, we set out to encode the underlying depth prior provided by our efficient ray sampling scheme by concatenating $\alpha$ of each sampling point estimated by Eqn.\ref{eq:alpha} together, i.e.,  $F_d = [\alpha_1, \alpha_1, \cdots, \alpha_{N}]$.


After that, we can obtain view-dependent radiance feature map $\mathcal{F}_c$ and density feature map $\mathcal{F}_d$ for each patch for the following convolutional rendering. 
Our novel patch-wise volumetric integration scheme provides high-level appearance and opacity features that enable high-quality RGB and alpha image synthesis.

\subsection{Convolutional Volume Renderer} \label{sec:Convolutional Volume Renderer}

Based on the above per-patch features which encode the characteristic of appearance and opacity, we introduce a convolutional volume renderer scheme for view-consistent appearance and alpha matte rendering by utilizing the spatial information, to address the issue that the original NeRF~\cite{martin2020nerf} fails to recover the high-frequency details of fuzzy objects with a simple MLP. 
Specifically, let $\bf G$ denote the network of our convolutional volume renderer with parameters $\theta$, which predicts the texture image $\bf{F}$ with corresponding opacity map $\alpha$ via $(\bf{F} , \alpha) = G(\mathcal{F}_c, \mathcal{F}_d; \theta)$, as illustrated in Fig.~\ref{fig:pipeline}. Different with NeRF which renders radiance pre-multiplied by alpha, $\bf F$ is just foreground image to avoid retaining radiance from the background near boundary. 

Note that we aim to achieve view consistent rendering, however, the local receptive field of convolutional neural network (CNN) itself may introduce view-inconsistency.
Thus $\bf G$ is designed to be light-weight to replace the last fully connect layer of the original NeRF network and decode the feature maps into radiance and opacity. In this way, the global implicit representation described in Sec.~\ref{sec:Feature_patch} plays a dominant role for view-consistency.  
%
%

As illustrated in Fig.~\ref{fig:pipeline}, our volume renderer $\bf G$ consists of a radiance branch and an opacity branch with similar U-Net architectures.
Note that in our radiance branch two downsample-upsample blocks are adopted to generate fine-detailed texture output, while only a single downsample-upsample block is utilized in our opacity branch.  
Such a sophisticated design is based on our observation that alpha mattes are sensitive to low-level features such as image gradients and are more suitable for a shallow network to preserve view-consistent output.

Besides, since the input feature maps $\mathcal{F}_c$ and $\mathcal{F}_d$ suffers from incomplete integration near the boundary regions, which is critical for fuzzy objects, we further adopt the gated convolution~\cite{yu2019free} in the U-Net architectures for both branches, so as to enhance the denoising and image completion capabilities of the network.
For more detailed alpha matte prediction, the texture output from the radiance branch is further concatenated with density feature map $\mathcal{F}_d$ to form a hybrid input of the opacity branch.
Also, we accumulate the $\mathcal{F}_d$ per-pixel first to form a coarse initial alpha matte, and the opacity branch predicts the opacity residuals.
Finally, the coarse matte and the residual one are added to generate the final detailed alpha matte output. 

To better encourage surface details (hairline, hair texture) and sharpness, we propose to apply a GAN-based discriminator loss to the final image patch (as shown in the right side of Fig.~\ref{fig:pipeline}). 
To enable self-supervised adversarial learning, we randomly sample patches from all the input multi-view images as real samples of the discriminator, and the fake samples are generated using the above volumetric render and the estimated alpha matte, which is formulated as:
\begin{equation}
    \bf{I}_{fake} = \alpha \bf{F} + (1 - \alpha) \bf{B},
\end{equation}
Where $\bf{F}, \bf{B}$ denote the generated foreground texture and background image, respectively
We set $B=[1.0,1.0,1.0]$ as a white background in all patches in our experiment.


\subsection{Network Training} \label{sec:Training}

We train the convolutional volume renderer $\bf G$ with $3$ generation loss and one discriminator loss. Recall that our model targets on high quality free-viewpoint radiance rendering (RGB and its opacity) from given sparse views RGBA images, we decomposite the prediction procedure into two layers (the RGB and alpha layers) and final compose them together via the predicted alpha layer. We first propose to use $L_2$ loss for the above two layers: 

\begin{equation}
\begin{aligned}
    \mathcal{L}_{C,\alpha} = \sum_{i=1}^{n} \|{\bf{I}}_i - {\tilde {\bf I}}_i\|_2^2 + \|{\bf{\alpha}}_i - {\tilde {\bf \alpha}}_i\|_2^2
\end{aligned}
\end{equation}
where $\bf \tilde I$ and $\tilde \alpha$ indicate the ground truth image and alpha. 

To encourage fine details and let outputs close to ground truth patches at the high level, We use a \textit{VGG19}  perceptual loss~\cite{johnson2016perceptual} on the output feature maps of the $l$th layer of the \textit{VGG19} backbone, which is denoted as $\phi^l$, over both the compositional image and alpha map: 
\begin{equation}
    \mathcal{L}_{P} = \sum_{l\in\{3,8\}}\sum_{i=1}^n(\|\phi^l_i(\textbf{I}) - \phi^l_i({\tilde{\textbf{I}}})\|_2^2 + \|\phi^l_i(\alpha) - \phi^l_i({\tilde{\alpha}})\|_2^2) 
\end{equation}

Instead of computing losses on the final outputs only, we add an intermediate loss to the MLP output to encourage multi-view consistency by considering the regional patch-based CNN renderer has smaller perceptual fields and can not capture global geometric, while the MLP block taking 3D location and view direction as input and adding constraint on its output can better preserve global consistency:
\begin{equation}
    \mathcal{L}_{N} = \sum_{i=1}^{n} (w_\alpha\|\alpha_i - \tilde{\alpha}_i\|_2^2 + w_i\|\textbf{I}_i - \tilde{\textbf{I}}_i\|_2^2 + \sum_{l\in\{3,8\}}\|\phi^l_i(\textbf{I}) - \phi^l_i({\tilde{\textbf{I}}})\|_2^2)
\end{equation}
where $w_\alpha = 1$ if the scene is synthetic otherwise $0$ as the ground truth alpha map of real objects is not view consistent,  that is, we only implicitly supervise the alpha with $RGB$ by considering stumpy gt alpha would result in jitter phenomenon on the boundary region.  Note that, $w_i = a - b \alpha_i$  where $a - b = 1$, this term penalties the false prediction on empty region.   We set $a=2, b=1 $ in our experiment. 

In summary, our optimizing objective function for generator $\mathcal{L}_G$ is:
\begin{equation}
    \mathcal{L}_G = \mathcal{L}_{C, \alpha} + \mathcal{L}_P +\mathcal{L}_N
\end{equation}

For the discriminator $D$, we minimize the adversarial loss~\cite{isola2017image}:
\begin{equation}
    \mathcal{L}_{D} = \|D(\textbf{I})\|_2^2 + \|D(\tilde{\textbf{I}}) - 1\|_2^2
\label{eq:adversarial loss}
\end{equation}




\begin{figure*}[!t]
	\centering
	\includegraphics[width=0.95\linewidth]{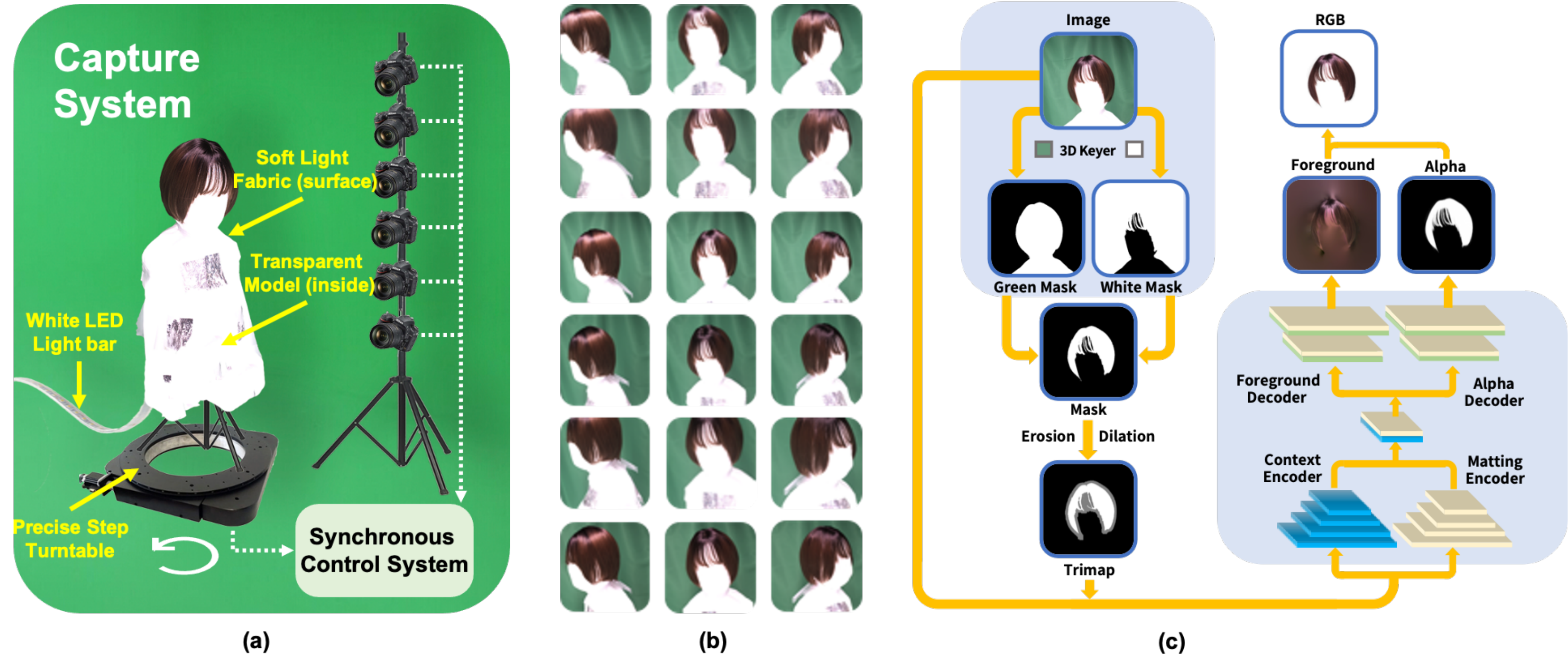}
	\vspace{-15pt}
	\caption{(a) Our Capture system consists of a lit support, an array of calibrated cameras, a precise step turntable, and a green screen. The fuzzy object is placed on the lit support and the precise turntable is synchronized with the camera array. (b) Samples of images captured among 6 cameras in 3 different steps. (c) The pipeline of alpha matte generation. }
	\label{fig:capture system}
\end{figure*}

\section{Capture System} \label{sec:system}
Here, we describe our capture system, which produces high-quality multi-view RGBA (RGB and Alpha) images for explicit opacity modeling of challenging fuzzy objects.
As illustrated in Fig.~\ref{fig:capture system}, our pipeline is equipped with an easy-to-use capture device and stable calibration and automatic matting methods. 

\myparagraph{Device.} 
Opacity object data could be captured through well-designed acquisition systems proposed in NOPC~\cite{wang2020neural}. 
However, there are some limitations, including disability in capturing objects without support (e.g., hair without a head model) or capturing without the calibration box. 
Thus we build a novel capture system as shown in Fig.~\ref{fig:capture system}(a), which consists of a precisely controllable step turntable placed in front of the green screen and six Nikon D750 DSLR Cameras facing towards the objects. 
For wig data, we use a transparent support covered with soft light fabric to support the object and place them on the turntable. Insert an LED light bar into the support to make the soft light fabric lit. Before capturing, the cameras are adjusted to overexpose the soft fabric part, which will reduce the shadows of the captured image. 
%
%
When capturing images, the turntable moves forward in precise steps. The turntable will step forward 80 steps per lap. All six cameras capture an image per step, which will serve as the training data.


\myparagraph{Calibration.}
Unlike general 3D object reconstruction, fuzzy objects are difficult to reconstruct using structure-from-motion(SFM) techniques, which means calibration via the reconstruction process would not work well. 
Previous work\cite{wang2020neural} use an auxiliary calibration camera with a pattern box to solve this, which takes much time to run an extra SFM pipeline for every capturing. 
In our system, the calibration process only needs one time for each camera setting. We know exactly how the camera's external parameters are transformed from the initial step to each step via the precise step turntable. Let $\mathbf{A}_j$ denote the affine transformation from the initial step to the $j$th step. 
The calibration process is divided into two steps. Firstly, calibrate the intrinsic and extrinsic parameters of six cameras at the initial step via Zhang’s camera calibration\cite{zhang1999flexible}, denote the $i$th camera’s intrinsic parameter as $\mathbf{K}_i$  and extrinsic parameter at its $j$th step as $\mathbf{T}_{i,j}$. Then, calculate the extrinsic parameter of each view under the turntable's coordinate

\begin{equation}
\mathbf{T}_{i,j} =\mathbf{A}_{j}\mathbf{T}_{i,0}
\end{equation}
where $i$ is the index of cameras, and $j$ is the number of steps.

\myparagraph{Opacity Decomposition.}
Image matting is an ill-posed problem due to a lack of constraint in its formulation. 
To obtain alpha matte without loss of rich details from fuzzy objects, we apply both 3D keyer and deep learning-based matting algorithm (see Fig.~\ref{fig:capture system}(c)). 
Specifically, the keying benefits from our specially designed capture system. In addition to the traditional green screen, the overexposure of support reduces the shadow cast by the object, providing a clean white screen. We key out both green and white from the image to extract a preliminary foreground mask. A trimap is then generated from the mask using erosion and dilation operations. We apply Context-aware Matting~\cite{hou2019context} to predict alpha matte and foreground simultaneously,
which combines the benefits of local propagation and global contextual information and performs visually much better on intractable data like curly hair and air bangs.

\begin{figure*}[htp]
\centering
  \includegraphics[width=1.0\linewidth]{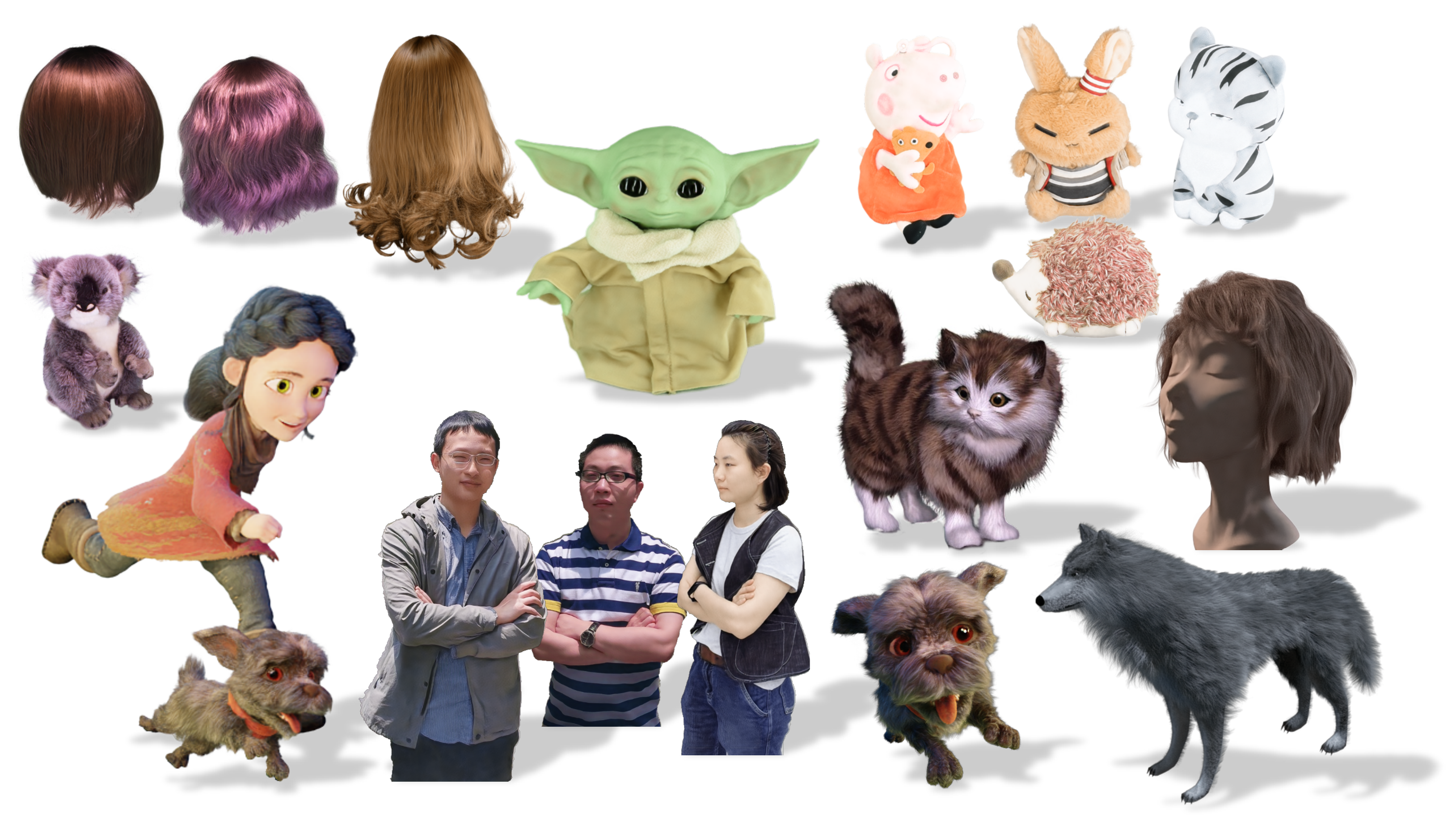}
  \caption{Object gallery. Our method can generalize well to various fuzzy objects, including high frequency, view-dependency and translucency appearance, such as hairstyles, clothes, toys and animals etc.}
  \label{fig:gallery}
\end{figure*}

\begin{figure*}[t]
\centering
  \includegraphics[width=0.95\linewidth]{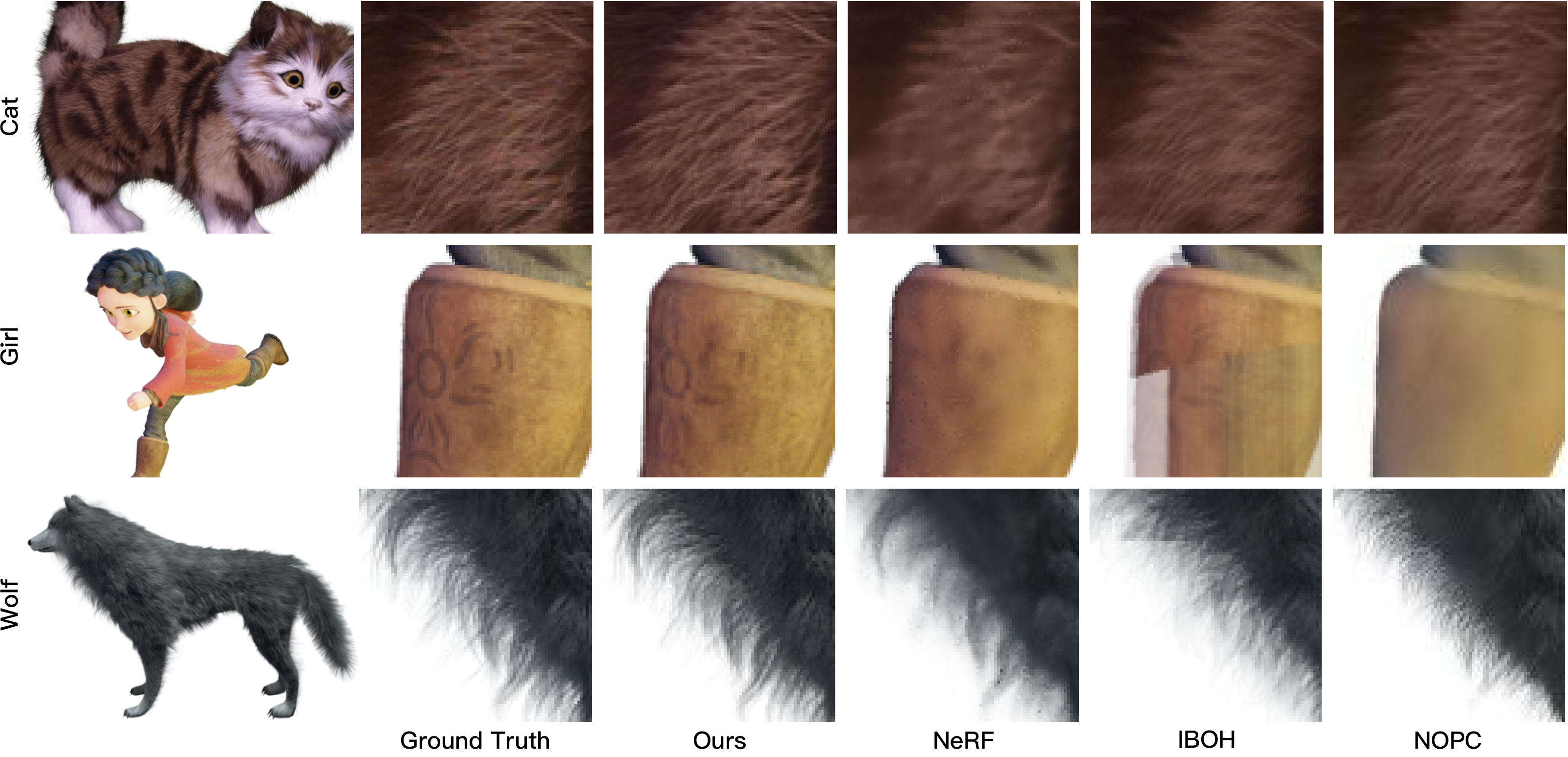}
  \caption{Free Viewpoint RGB results comparison with IBOH ~\cite{matusik2002image}, NOPC~\cite{wang2020neural} and NeRF ~\cite{mildenhall2020nerf} on \textit{Cat}, \textit{Girl}, \textit{Wolf} datasets. Our method is able to reconstruct fine details on both geometry and appearance while keeping global view-consistency, such as \textit{Cat}'s fur texture, the pattern on \textit{Girl}'s boots and the geometric details of wolf's hair. IBOH exhibits ghosting and aliasing, NOPC exhibits excessive blurs and loss of geometric details and NeRF exhibits excessive noise and blurs. See supplementary materials for more results.}
  \label{fig:rgb_comparison}
\end{figure*}

\section{Experiments}
\label{sec:experiment}

We evaluate our ConvNeRF on various furry objects. Quantitative and qualitative evaluation in Sec.~\ref{sec:comparison} show that our method can better preserve high fidelity appearance details than prior work and generate globally consistent alpha mattes in arbitrary novel views. We further perform extensive ablation studies to validate our design choices in Sec.~\ref{sec:ablation}. We urge the reader to view our supplementary video to better appreciate our method’s significant improvement over baseline methods when rendering novel views. 


\myparagraph{Baselines.} We adopt the conventional IBR method \textit{Image-based Opacity Hull (IBOH)~\cite{matusik2002image}} and recent explicit neural rendering method \textit{Neural Opacity Point Cloud (NOPC)~\cite{wang2020neural}} and implicit volume rendering method \textit{Neural Radiance Fields (NeRF)~\cite{mildenhall2020nerf}} as baselines for comparisons. 

\myparagraph{Training details.} We use half original image resolution for synthetic and real objects, $200\sim300$ images for training. We use patches of size $32\times 32$  and 64 samples (32 in coarse and fine modules) per ray for ConvNeRF training. For the baseline models, we sample 300,000 points for each object from the same proxy mesh with us as the dense point cloud input for the \textit{NOPC}. Also, we use 128 samples (64 in coarse and fine modules) for \textit{NeRF} training. It takes $1\sim 2$ days per object to train our model with a single NVIDIA TITAN RTX GPU, while the \textit{NeRF} takes about a  week if trained on the same image number and resolution. 

\subsection{Comparison}
\label{sec:comparison}

\begin{figure*}[!t]
\centering
  \includegraphics[width=0.95\linewidth]{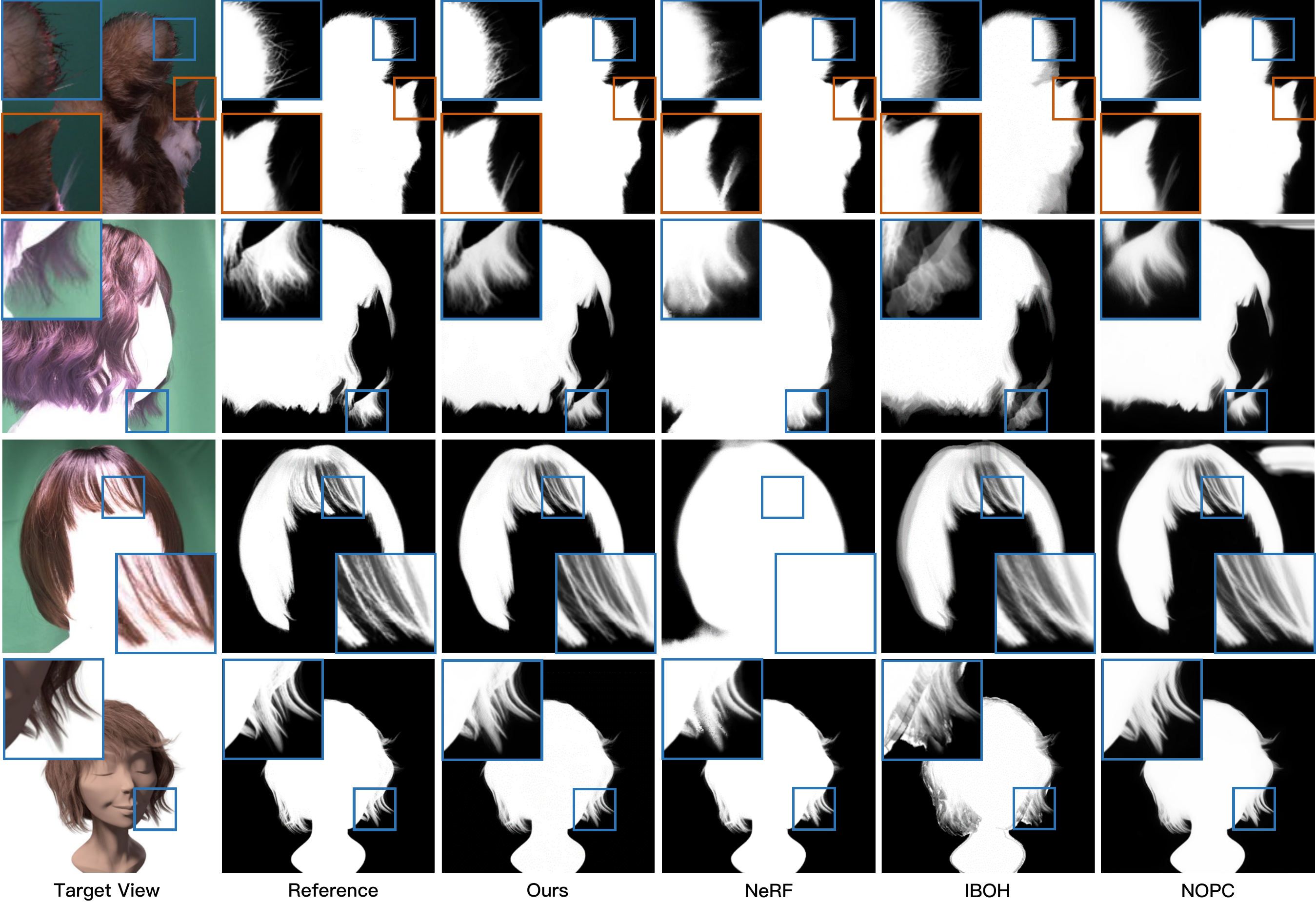}
  \caption{Free Viewpoint Alpha results comparison with IBOH ~\cite{matusik2002image}, NOPC~\cite{wang2020neural} and NeRF ~\cite{mildenhall2020nerf} on \textit{Cat}, \textit{Hairstyle 2} datasets. Our method can recover missing part opacity such \textit{Cat}'s beard from view-inconsistent alpha mattes as shown in the first row, While IBOH fails and exhibits strong polygonal artifacts. Our method can produce sharper alpha mattes than NOPC which suffers from strong artifacts around the hair. NeRF fails on our challenging \textit{Hairstyle 2} dataset. See supplementary materials for more results. }
  \label{fig:alpha_comparison}
\end{figure*}

We perform quantitative and qualitative comparisons on both RGB and alpha with the above baseline models. For qualitative evaluation, Fig.~\ref{fig:rgb_comparison} shows several our novel viewpoint RGB rendering results together with visual compassion v.s. most recent neural scene representation methods. We can see that the \textit{IBOH} suffers from ghosting and aliasing due to the input views are in-uniform sampled. The point cloud based \textit{NOPC} can partially generate smooth texture details with sufficient training images but still suffers from color shifting and blur on the \textit{Girl} object due to interpolation between the features projected from the discrete spacial point cloud. The NeRF can preserve low-frequency components but fails to recover high-frequency details due to the limited representation ability of the MLP network. In the fur region, there is noise caused by insufficient samples on both ray direction and the ray resolutions. In contrast, our ConvNeRF is able to reconstruct fine texture and geometry details and present sharply visual results with global consistency (refer to our supplemental video).  
 
Fig.~\ref{fig:alpha_comparison} shows the visual discrepancies of free-viewpoint alpha maps rendering results from given discrete view samples. Compared to other methods, our ConvNeRF is able to preserve the sharpness of the hair boundary. Note that we can easily obtain the perfect alpha map in synthesis scenes while not easy for real data; our method is not sensitive to the quality of input alpha maps and is able to partially recover missing part from nearby view samples (as shown in the first row of Fig.~\ref{fig:alpha_comparison}).

We quantitatively evaluate our method with PSNR, LPIPS (ALEX backbone) ~\cite{zhang2018perceptual} and SSIM metric. \footnote{We computer quantity value only on foreground region (i.e., the object itself) by considering the background is not included in the final render.} As shown in Tab.~\ref{table:rgb_comparison} and Tab.~\ref{table:alpha_comparison}, our ConvNeRF achieves significant improvement on both RGB and alpha results. We replace the LPIPS metric with \textit{Sum of Absolute Distance} (SAD) when evaluating the alpha result due to the domain gap between the alpha-like image and the training set of the network-based distance metric. Tab.~\ref{table:average_metric} shows the average PSNR of all datasets on semi-translucent region (i.e., $0 < \alpha < 1$). Our method achieves state-of-the-art performance.

\begin{table}[!t]
\tiny
\centering
\caption{Quantitative comparisons of foreground images on different Objects}
\resizebox{1\linewidth}{!}{
\begin{tabular}{m{6em}<{\centering} m{4em}<{\centering} c c c c} 
\toprule
Method & & IBOH & NOPC & NeRF & Ours \\ [0.5ex] 
\midrule
\multirow{3}{*}{\bf Wolf}           & PSNR  & 28.25 & 29.95 & 32.56 & \textbf{37.23} \\ 
                                    & SSIM  & 0.932 & 0.945 & 0.965 & \textbf{0.975} \\
                                    & LPIPS & 0.073 & 0.058 & 0.061 & \textbf{0.022} \\
\midrule
\multirow{3}{*}{\bf Hair}           & PSNR  & 23.22 & 28.31 & 28.29 & \textbf{33.56} \\ 
                                    & SSIM  & 0.872 & 0.912 & 0.932 & \textbf{0.952} \\
                                    & LPIPS & 0.104 & 0.058 & 0.070 & \textbf{0.021} \\
\midrule
\multirow{3}{*}{\bf Girl}           & PSNR  & 18.32 & 26.97 & 29.77 & \textbf{32.17} \\ 
                                    & SSIM  & 0.863 & 0.922 & 0.951 & \textbf{0.958} \\
                                    & LPIPS & 0.150 & 0.099 & 0.052 & \textbf{0.035} \\
\midrule
\multirow{3}{*}{\bf Dog}            & PSNR  & 22.53 & 27.88 & 29.59 & \textbf{31.87} \\ 
                                    & SSIM  & 0.866 & 0.902 & 0.937 & \textbf{0.950} \\
                                    & LPIPS & 0.127 & 0.098 & 0.053 & \textbf{0.030} \\
\midrule
\multirow{3}{*}{\bf Koala}          & PSNR  & 24.41 & 25.45 & 32.90 & \textbf{33.48} \\ 
                                    & SSIM  & 0.871 & 0.890 & 0.960 & \textbf{0.965} \\
                                    & LPIPS & 0.103 & 0.117 & 0.053 & \textbf{0.023} \\
\midrule
\multirow{3}{*}{\bf Cat}            & PSNR  & 20.40 & 26.72 & 24.80 & \textbf{28.95} \\ 
                                    & SSIM  & 0.810 & 0.888 & \textbf{0.896} & \textbf{0.894} \\
                                    & LPIPS & 0.198 & 0.101 & 0.187 & \textbf{0.090} \\
\midrule
\multirow{3}{*}{\bf Hairstyle 1}    & PSNR  & 25.32 & 27.34 & 17.84 & \textbf{30.51} \\ 
                                    & SSIM  & 0.905 & 0.921 & 0.904 & \textbf{0.949} \\
                                    & LPIPS & 0.067 & 0.058 & 0.135 & \textbf{0.037} \\
\midrule
\multirow{3}{*}{\bf Hairstyle 2}    & PSNR  & 25.86 & 31.14 & 19.38 & \textbf{37.21} \\ 
                                    & SSIM  & 0.951 & 0.962 & 0.931 & \textbf{0.983} \\
                                    & LPIPS & 0.104 & 0.033 & 0.131 & \textbf{0.014} \\
\midrule
\multirow{3}{*}{\bf Hairstyle 3}    & PSNR  & 14.26 & 29.38 & 22.69 & \textbf{33.65} \\ 
                                    & SSIM  & 0.877 & 0.942 & 0.944 & \textbf{0.969} \\
                                    & LPIPS & 0.130 & 0.040 & 0.073 & \textbf{0.016} \\
\bottomrule
\end{tabular}
}
\label{table:rgb_comparison}
\end{table}

\begin{table}[!t]
\tiny
\centering
\caption{Quantitative comparisons of alpha mattes on different Objects}
\resizebox{1\linewidth}{!}{
\begin{tabular}{m{6em}<{\centering} m{4em}<{\centering} c c c c} 
\toprule
Method & & IBOH & NOPC & NeRF & Ours \\ [0.5ex] 
\midrule
\multirow{3}{*}{\bf Wolf}           & SAD  & 18.46 & 7.977 & 4.777 & \textbf{2.564} \\ 
                                    & PSNR & 25.10 & 29.03 & 32.14 & \textbf{38.63} \\
                                    & SSIM & 0.956 & 0.983 & 0.990 & \textbf{0.996} \\
\midrule
\multirow{3}{*}{\bf Hair}           & SAD  & 18.93 & 6.117 & 5.078 & \textbf{2.077} \\ 
                                    & PSNR & 21.87 & 29.54 & 28.25 & \textbf{36.35} \\
                                    & SSIM & 0.930 & 0.981 & 0.981 & \textbf{0.994} \\
\midrule
\multirow{3}{*}{\bf Girl}           & SAD  & 48.61 & 12.66 & 5.290 & \textbf{2.877} \\ 
                                    & PSNR & 19.59 & 27.83 & 30.58 & \textbf{36.39} \\
                                    & SSIM & 0.937 & 0.986 & 0.991 & \textbf{0.996} \\
\midrule
\multirow{3}{*}{\bf Dog}            & SAD  & 30.77 & 11.32 & 6.828 & \textbf{3.661} \\ 
                                    & PSNR & 22.77 & 27.78 & 30.41 & \textbf{36.52} \\
                                    & SSIM & 0.945 & 0.983 & 0.988 & \textbf{0.995} \\
\midrule
\multirow{3}{*}{\bf Koala}          & SAD  & 26.04 & 71.32 & 190.15 & \textbf{18.94} \\ 
                                    & PSNR & 22.04 & 21.76 & 14.56 & \textbf{30.26} \\
                                    & SSIM & 0.939 & 0.962 & 0.904 & \textbf{0.983} \\
\midrule
\multirow{3}{*}{\bf Cat}            & SAD  & 50.85 & 21.46 & \textbf{20.81} & 23.79 \\ 
                                    & PSNR & 19.17 & \textbf{25.67} & 24.25 & \textbf{25.52} \\
                                    & SSIM & 0.913 & 0.958 & \textbf{0.960} & 0.944 \\
\midrule
\multirow{3}{*}{\bf Hairstyle 1}    & SAD  & 15.51 & 9.761 & 30.66 & \textbf{5.445} \\ 
                                    & PSNR & 25.06 & 28.96 & 18.16 & \textbf{32.48} \\
                                    & SSIM & 0.961 & 0.977 & 0.947 & \textbf{0.987} \\
\midrule
\multirow{3}{*}{\bf Hairstyle 2}    & SAD  & 7.677 & 5.948 & 22.69 & \textbf{2.612} \\ 
                                    & PSNR & 28.80 & 32.01 & 19.46 & \textbf{37.54} \\
                                    & SSIM & 0.980 & 0.988 & 0.957 & \textbf{0.994} \\
\midrule
\multirow{3}{*}{\bf Hairstyle 3}    & SAD  & 46.09 & 10.06 & 12.58 & \textbf{3.037} \\ 
                                    & PSNR & 24.13 & 26.20 & 22.77 & \textbf{34.62} \\
                                    & SSIM & 0.960 & 0.978 & 0.973 & \textbf{0.993} \\
\bottomrule
\end{tabular}
}
\label{table:alpha_comparison}
\end{table}


\begin{table}[!t]
\centering
\caption{Average PSNR on semi-translucent region (U). U+ and U- represent the region after dilation and erosion. }
\begin{tabular}{cccccccc}
\toprule
\multicolumn{2}{c}{\multirow{2}{*}{Method}} & \multicolumn{3}{c}{$\alpha \mathbf{F}$} & \multicolumn{3}{c}{Alpha} \\ \cline{3-8}
\specialrule{0em}{1pt}{1pt}
\multicolumn{2}{c}{}                        &    U-    &    U   &    U+   &     U-    &    U    &    U+    \\
\midrule
\multicolumn{2}{c}{IBOH}                    &    15.78    &    16.05  &   16.46    &    12.93     &    12.99    &  13.59     \\
\multicolumn{2}{c}{NOPC}                    &    16.24    &    17.49  &   18.75    &    14.15     &    15.14    &  16.55     \\
\multicolumn{2}{c}{NeRF}                    &    12.95    &    13.09  &   14.83    &    12.17     &    12.48    &  14.29     \\
\multicolumn{2}{c}{Ours}                    & \textbf{25.23} & \textbf{26.21} & \textbf{27.06} & \textbf{20.73} & \textbf{22.45}    & \textbf{24.21}      \\
\bottomrule
\end{tabular}
\label{table:average_metric}
\end{table}



\subsection{Evaluation}
\label{sec:ablation}
To explore each part's contribution in our pipeline, we conduct ablation studies by choosing a set of controlled experiments, including randomized ray sampling v.s. efficient ray sampling, Pixel-wised v.s. Convolutional renderer, with/without GAN based discriminator loss, the range of the sampling Number and patch size. 

\begin{figure*}[t!]
\centering
  \includegraphics[width=1.0\linewidth]{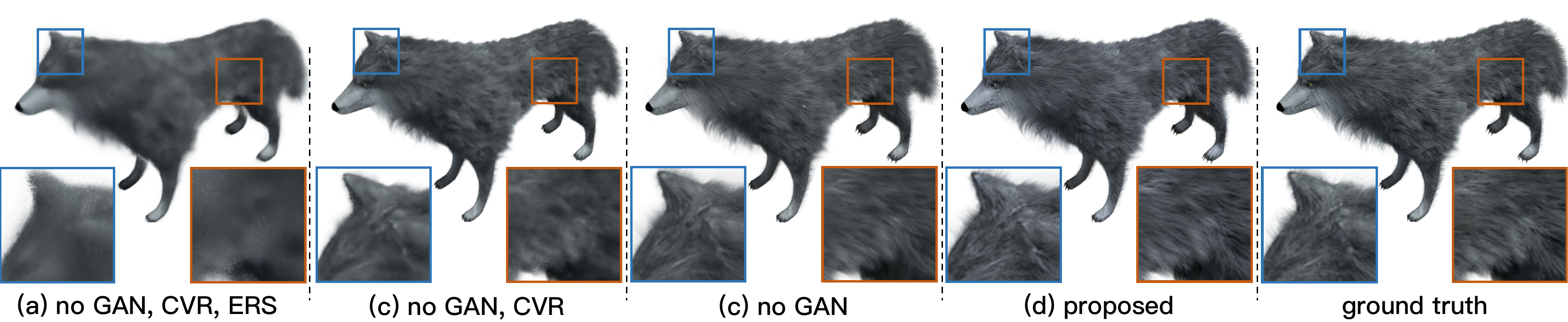}
  \caption{Visualization of the ablation study on w/o the ERS, CVR and GAN loss modules (corresponding to the Tab \ref{table:ablation}). }
  \label{fig:ablation}
\end{figure*}

\myparagraph{Randomize v.s. Efficient Ray Sampling} We first compare the proposed Efficient Ray Sampling (ERS) scheme described in Sec.~\ref{sec:BSampling}, which include a baseline model (a) (i.e., \textit{NeRF}) with 64 samples per ray same as our ConvNeRF and a model (b) with ERS. As the \textit{Wolf} case shown in Fig.~\ref{fig:ablation}, model (a) fails to recover surface details while ERS brings significant improvements to the visual quality and quantitative quality shown by model (a) and (b) in Tab.~\ref{table:ablation}.

\myparagraph{Pixel-wised v.s. Convolutional Volume Renderer} We train a generator model without the discriminator, as (c) denoted in Tab.~\ref{table:ablation} and Fig.~\ref{fig:ablation}, setting (c) can recover better texture details on the wolf's forehead while (a) and (b) can only preserve low frequency information.  

\myparagraph{W/o GAN Loss} We further evaluate the effect of the GAN discriminator adopted for fine details refinement. The surface details of (c) are a little over-smoothed while (d) present fine and sharp details on the fur texture and hairlines, which is almost close to the ground truth. As for quantitative metrics, (d), setting keeps high performance on foreground texture and alpha quality, as the  (c) and (d) shown in Tab.~\ref{table:ablation}.


\begin{figure}[!t]
\centering
    \includegraphics[width=1.0\linewidth]{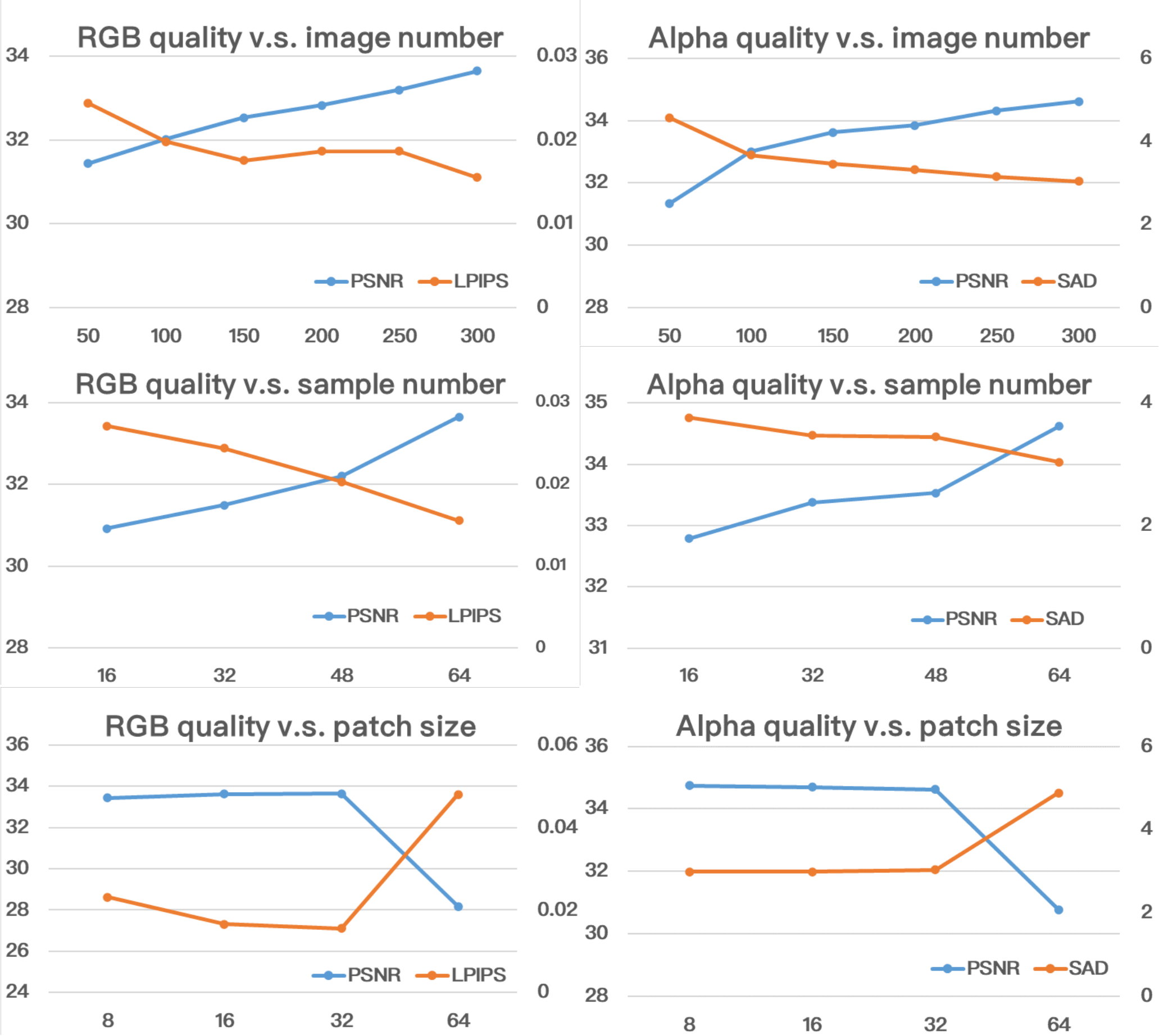}
\caption{From top to bottom: rendering quality v.s. training image number, samples per ray, patch size}
\label{fig:ablation_curve}
\end{figure}

\myparagraph{Number of Sample Points.} We further demonstrate the effect of different numbers of per ray sample points and image numbers. We evaluate on $N \in \{16, 32, 48, 64\}$ and image number $S \in \{50, 100, 150, 200, 250, 300\}$ respectively, as shown in the first and second row of Fig.~\ref{fig:ablation_curve}. More sample points can lead to better rendering quality and higher alpha prediction accuracy such as hairlines. Note that our method is not sensitive to training data size and can produce satisfactory results even with a few samples per ray. 

\begin{figure}[!t]
\centering
    \includegraphics[width=1.0\linewidth]{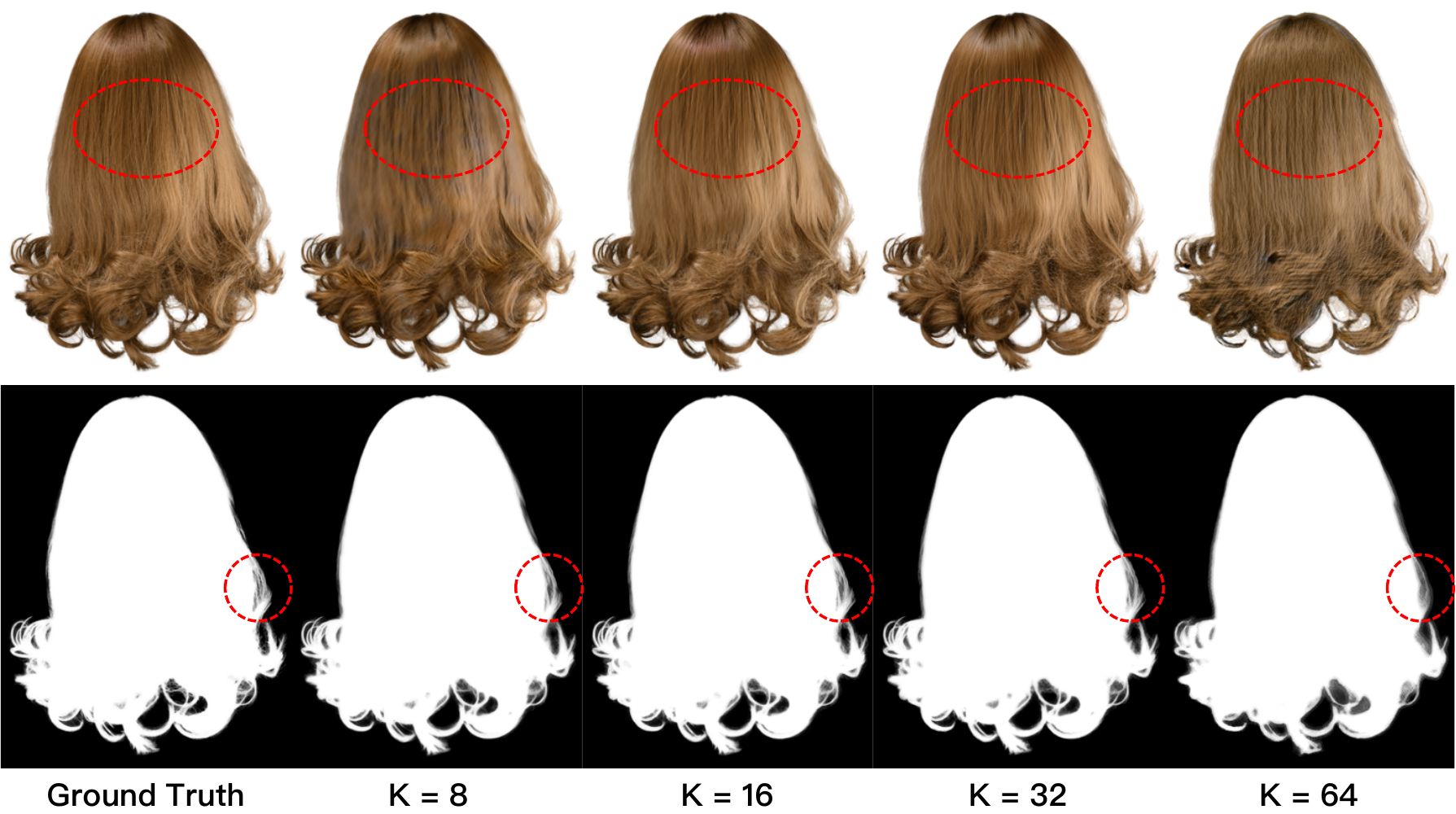}
\caption{Ablation study on the patch size of $K = [8, 16, 32, 64]$. Small and large sizes produce color cast and blurry artifacts. }
\label{fig:patch}
\end{figure}

\myparagraph{Patch Size.} As we introduce the convolutional mechanism in the image plane, the patch size plays a significant role for synthesis quality. Larger patch size helps preserve high-order features through perceptual loss while leads to fewer patches for training, which reduces data diversity for adversarial training. To evaluate the effect of different patch sizes, we set the patch size $K \in\{8, 16, 32, 64\}$ and conduct quantitative(Fig.~\ref{fig:ablation_curve}) and qualitative(Fig.~\ref{fig:patch}) comparisons. Notice that although the model with $K = 8$ achieves similar quantitative performance with our proposed model, it exhibits severe visual artifacts (the first row of Fig.~\ref{fig:patch}). 

\myparagraph{Textured Background.} As we mask out the background for real data, it may affect the rendering quality of both NeRF and our model. Thus we render the synthetic scene \textit{Wolf} together with rich texture backgrounds and parallax to evaluate our approach against NeRF. Tab.~\ref{table:background} indicates that both models perform slightly better in radiance rendering, while using textured background seems to downgrade the alpha rendering quality of both our approach and NeRF. It follows the insight that utilizing explicit alpha supervision is more effective than relying on the network itself to extract such alpha cues implicitly.

\begin{table}[t!]
\caption{Quantitative evaluation of the ablation studies.}
\begin{center}
\begin{tabular}{cccccccccc}
\toprule
\multicolumn{4}{c}{\multirow{2}{*}{Models}} & \multicolumn{2}{c}{$\alpha \mathbf{F}$} & \multicolumn{2}{c}{Alpha} \\\cline{5-8}
\specialrule{0em}{1pt}{1pt}
\multicolumn{4}{c}{}                        & PSNR $\uparrow$ & LPIPS $\downarrow$ & PSNR $\uparrow$ & SAD $\downarrow$ \\
\midrule

\multicolumn{4}{l}{(a) w/o GAN, CVR, ERS}       & 20.27 & 0.175 & 20.18 & 23.51 \\
\multicolumn{4}{l}{(b) w/o GAN, CVR}            & 31.38 & 0.074 & 34.46 & 3.808 \\
\multicolumn{4}{l}{(c) w/o GAN}                 & \textbf{37.55} & 0.026& 37.93 & 2.642 \\
\multicolumn{4}{l}{(d) ConvNeRF}                     & \textbf{37.23} & \textbf{0.022} & \textbf{38.63} & \textbf{2.564} \\

\bottomrule
\end{tabular}
\label{table:ablation}
\end{center}
\end{table}

\subsection{Limitation and Discussion}
\label{sec:limitation}
Our model is able to reconstruct high-frequency appearance from given multi-view RGBA images, the combination of our patch-wise learning and the global implicit representation described in Sec.~\ref{sec: algorithm} encourage view-consistency, however, we note a bit of flicker in the around-view videos. It is mainly due to the patch-wise adversarial loss in Eqn.~\ref{eq:adversarial loss} which increases high-frequency details significantly with a slight trade-off of view consistency.

On the other hand, our method relies on accurately calibrated camera pose; inaccurate camera poses for in the wild images will lead to loss of geometric details especially for fuzzy objects. Furthermore, since our method requires background images captured or separated by matting algorithms, it is almost impossible to manually capture the aligned background or obtain accurate alpha matte of images with a complex background, resulting in hard to extend it to in the wild data.

\begin{table}[t!]
\caption{Quantitative evaluation of textured background.}
\begin{center}
\begin{tabular}{cccccccccc}
\toprule
\multicolumn{4}{c}{\multirow{2}{*}{Background}} & & \multicolumn{2}{c}{$\alpha \mathbf{F}$} & \multicolumn{2}{c}{Alpha} \\\cline{6-9}
\specialrule{0em}{1pt}{1pt}
\multicolumn{4}{c}{} &  & PSNR $\uparrow$ & LPIPS $\downarrow$ & PSNR $\uparrow$ & SAD $\downarrow$ \\
\midrule

\multicolumn{4}{l}{\multirow{2}{*}{textured }}              & NeRF & 33.26          & 0.060          & 31.05          & 6.478 \\
\multicolumn{4}{c}{}                                        & Ours & \textbf{37.95} & 0.023          & 35.51          & 4.309 \\

\multicolumn{4}{l}{\multirow{2}{*}{white    }}              & NeRF & 32.56          & 0.061          & 32.14          & 4.777 \\
\multicolumn{4}{c}{}                                        & Ours & 37.23          & \textbf{0.022} & \textbf{38.63} & \textbf{2.564} \\

\bottomrule
\end{tabular}
\label{table:background}
\end{center}
\end{table}

\section{Conclusion}
We have presented a novel neural rendering framework to combine both explicit opacity modeling and convolutional mechanism into the neural radiance field, enabling high-quality, globally consistent and free-viewpoint appearance and opacity rendering for fuzzy objects.
Our efficient sampling strategy enables efficient radiance field sampling and learning in a patch-wise manner, while our novel volumetric integration generates per-patch hybrid features to reconstruct the view-consistent fine-detailed appearance and opacity output.
Our novel patch-wise adversarial training scheme further preserves the high-frequency appearance and opacity details for photo-realistic rendering in a self-supervised framework. 
Our experimental results demonstrate the effectiveness of the proposed convolutional neural opacity radiance field for high-quality appearance and opacity modeling.
We believe that our approach is a significant step to enable photo-realistic modeling and rendering for challenging fuzzy objects, with many potential applications in VR/AR like gaming, entertainment and immersive telepresence.

\ifpeerreview \else
\section*{Acknowledgments}
This work was supported by NSFC programs (61976138, 61977047), the National Key Research and Development Program (2018YFB2100500), STCSM (2015F0203-000-06) and SHMEC (2019-01-07-00-01-E00003).
\fi

\bibliographystyle{IEEEtran}
\bibliography{egbib}

\ifpeerreview \else



\begin{IEEEbiography}[{\includegraphics[width=1in,height=1.25in,clip,keepaspectratio]{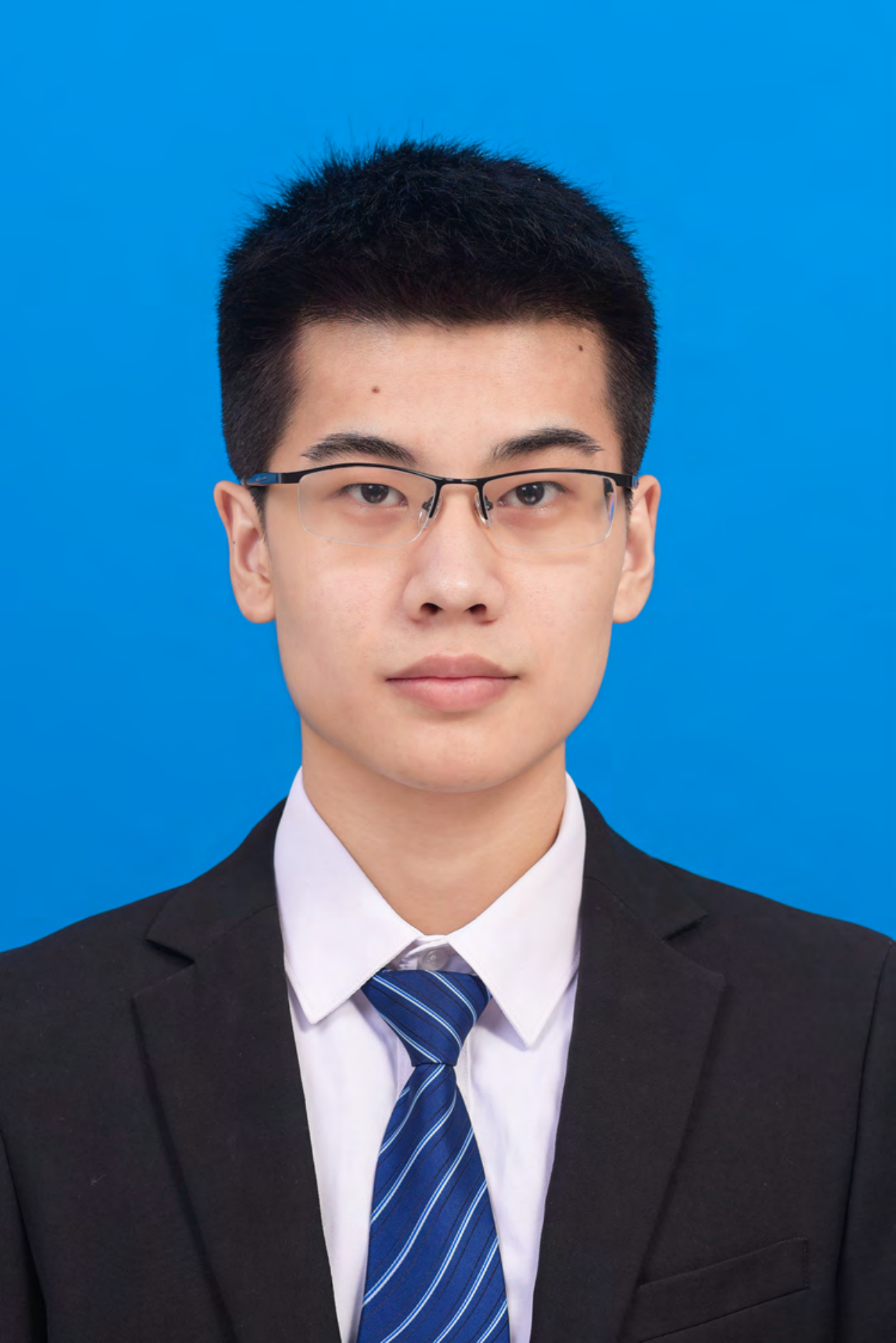}}]{Haimin Luo}
received the B.S. degree from the School of Computer Science and Technology, Huazhong University of Science and Technology, Wuhan, China, in 2019. He is currently working toward the master’s degree at ShanghaiTech University, Shanghai, China. His research interests include computer vision, computer graphics, and deep learning.

\end{IEEEbiography}

\begin{IEEEbiography}[{\includegraphics[width=1in,height=1.25in,clip,keepaspectratio]{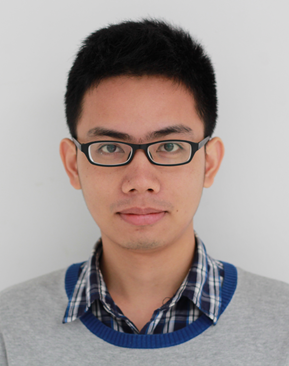}}]{Anpei Chen}
received the B.S. degree from the School of Physics and Optoelectronic Engineering, Xidian University, Shanxi, China, in 2016. He is currently working toward the PhD degree at ShanghaiTech University, Shanghai, China. His research interests lie at the intersection of computer graphics and computer vision, including image synthesis$/$editing, geometric modeling, and realistic rendering.
\end{IEEEbiography}

\begin{IEEEbiography}[{\includegraphics[width=1in,height=1.25in,clip,keepaspectratio]{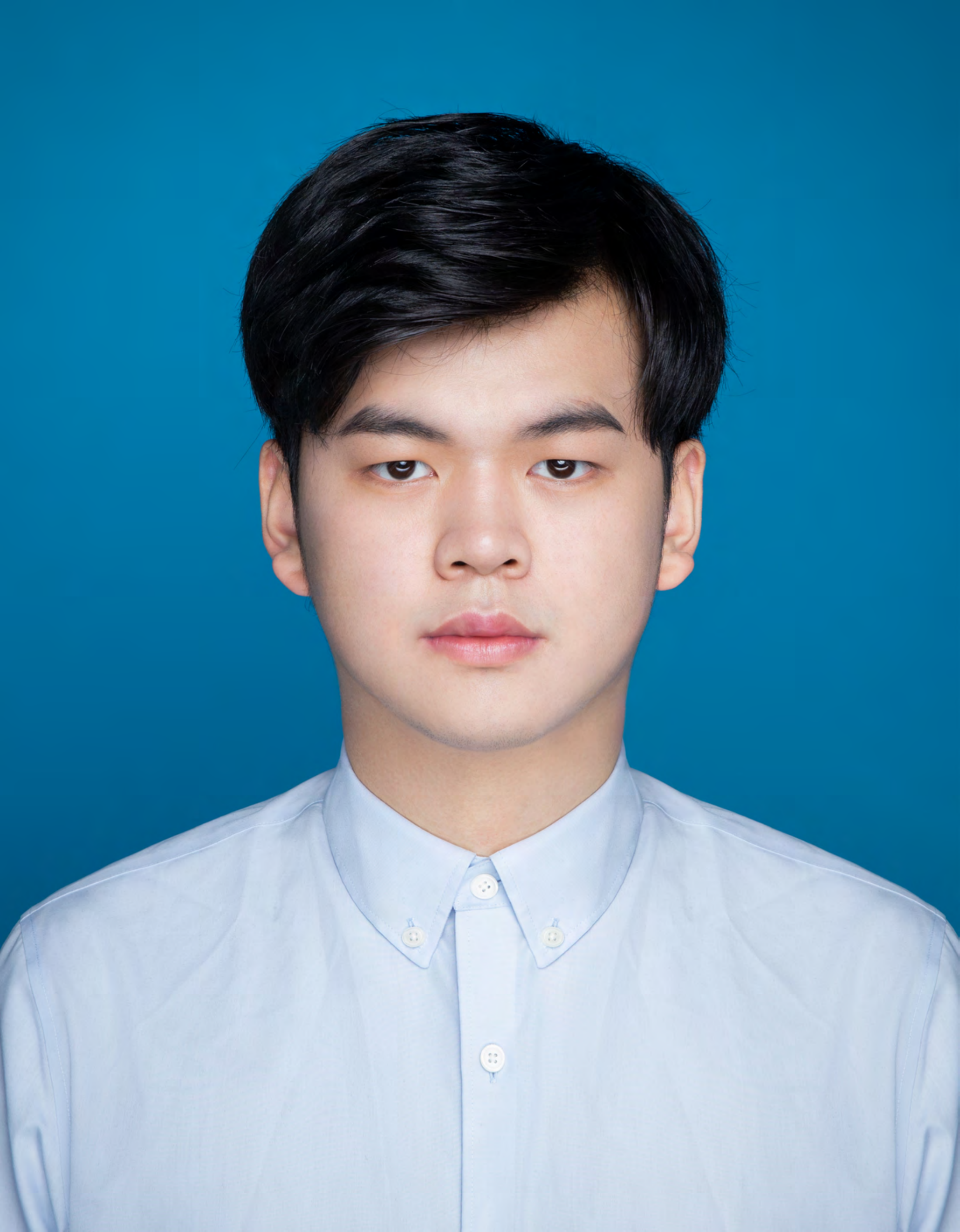}}]{Qixuan Zhang}
is currently working toward the B.S. degree at ShanghaiTech University, Shanghai, China. His research interests include computer vision, computational photography, and deep learning.
\end{IEEEbiography}

\begin{IEEEbiography}[{\includegraphics[width=1in,height=1.25in,clip,keepaspectratio]{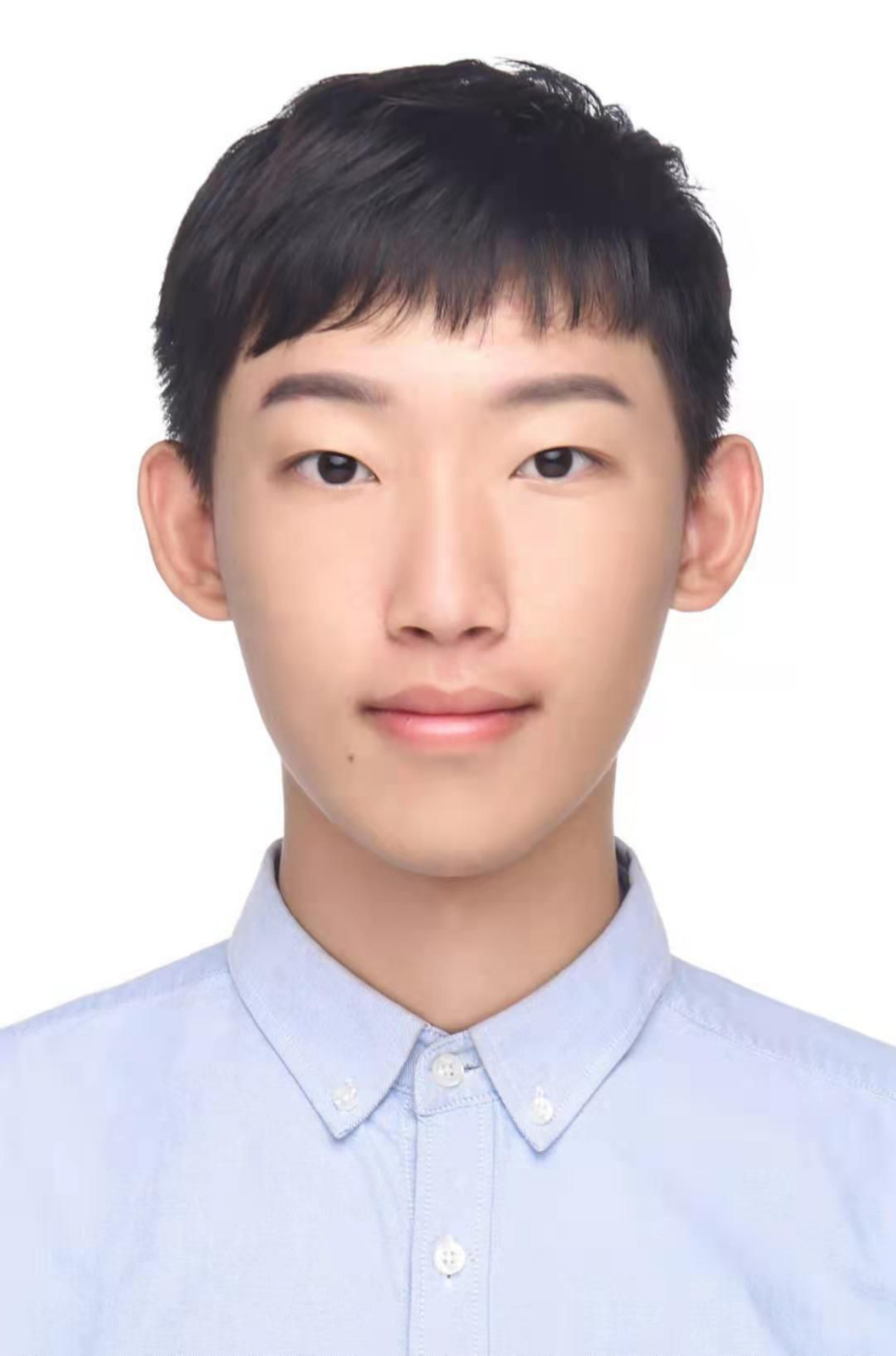}}]{Bai Pang}
is currently working toward the B.S. degree at ShanghaiTech University, Shanghai, China. His research interests include computer vision, computational photography and machine learning.
\end{IEEEbiography}

\begin{IEEEbiography}
[{\includegraphics[width=1in,height=1.25in,clip,keepaspectratio]{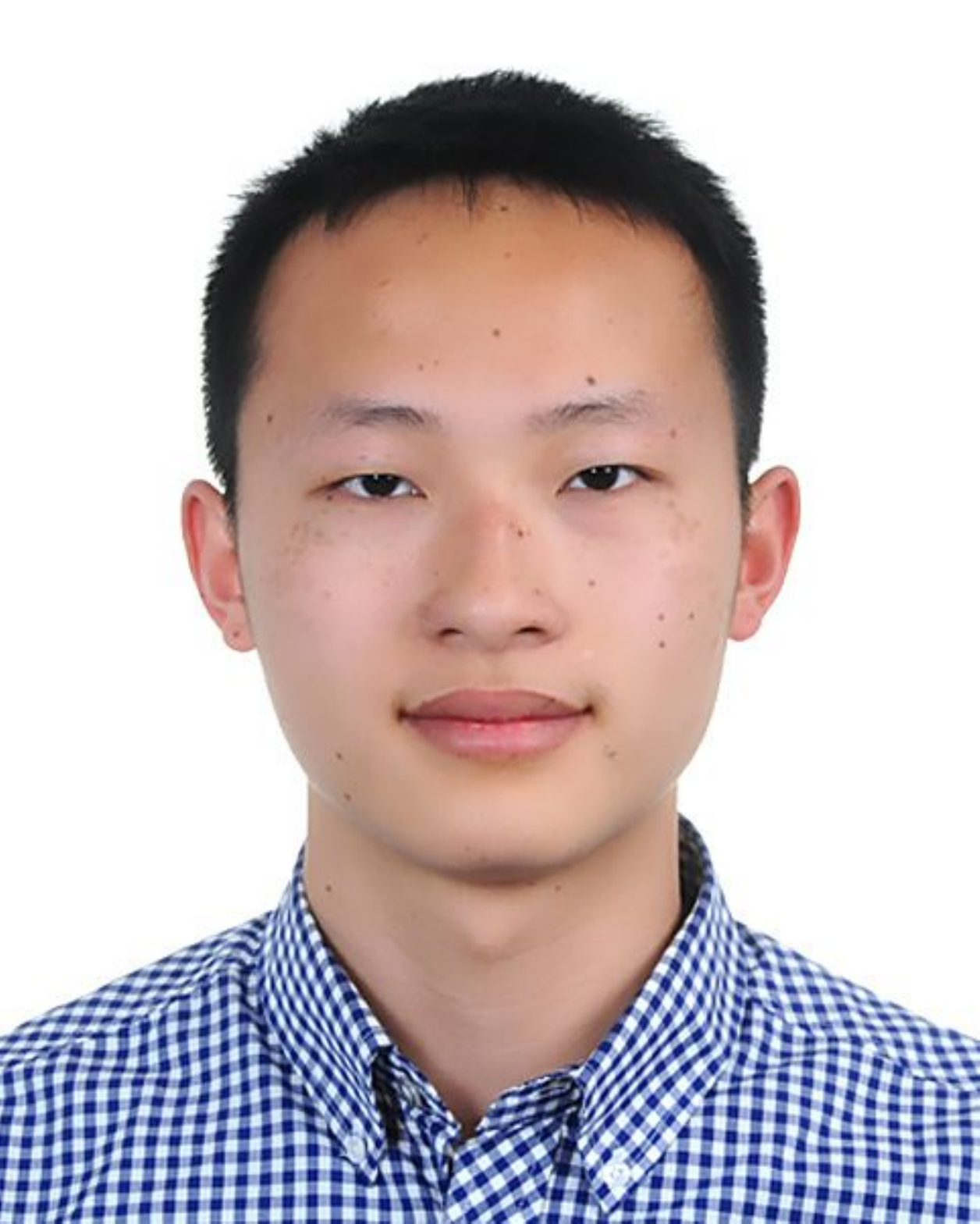}}]
{Minye Wu}
received the B.S. degree from the School of Computer Engineering and Science, Shanghai University, Shanghai, China, in 2015. He is currently working toward the PhD degree at ShanghaiTech University, Shanghai, China. He is also with the Shanghai Institute of Microsystem and Information Technology, Chinese Academy of Sciences, China, and the University of Chinese Academy of Sciences, China. His research interests include computer vision, deep learning, and computational photography.
\end{IEEEbiography}

\begin{IEEEbiography}
[{\includegraphics[width=1in,height=1.25in,clip,keepaspectratio]{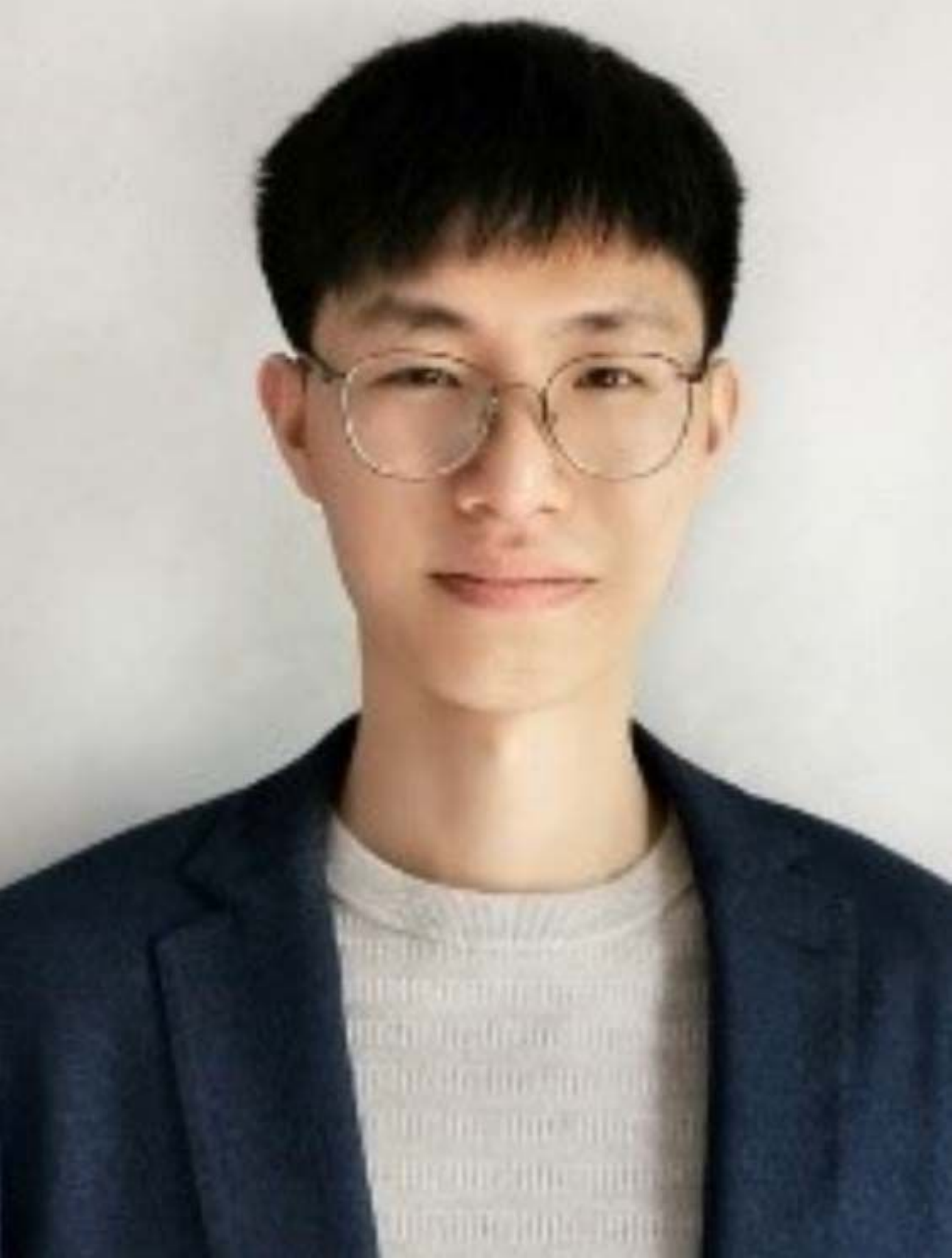}}]
{Lan Xu}
received the B.E. degree from Zhejiang University in 2015 and the Ph.D. degree from the Department of Electronic and Computer Engineering (ECE), The Hong Kong University of Science and Technology (HKUST), in 2020. He is currently an Assistant Professor with ShanghaiTech University. His research interests include computer vision, computer graphics, and machine learning.
\end{IEEEbiography}

\begin{IEEEbiography}
[{\includegraphics[width=1in,height=1.25in,clip,keepaspectratio]{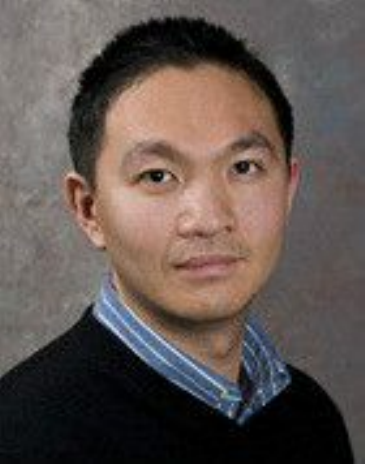}}]
{Jingyi Yu}
received B.S. from Caltech in 2000 and PhD from MIT in 2005. He is currently the Vice Provost at the ShanghaiTech University. Before joining ShanghaiTech, he was a full professor in the Department of Computer and Information Sciences at University of Delaware. His research interests span a range of topics in computer vision and computer graphics, especially on computational photography and non-conventional optics and camera designs. He is a recipient of the NSF CAREER Award and the AFOSR YIP Award, and has served as an area chair of many international conferences including CVPR, ICCV, ECCV, IJCAI and NeurIPS. He is currently a program chair of CVPR 2021 and will be a program chair of ICCV 2025. He has been an associate editor of the IEEE Transactions on Pattern Analysis and Machine Intelligence, the IEEE Transactions on Image Processing, and the Elsevier Computer Vision and Image Understanding. He is a fellow of IEEE. 
\end{IEEEbiography}



\vfill

\fi

\end{document}